\documentclass{article}

\PassOptionsToPackage{numbers,sort&compress}{natbib}
\usepackage[position,final]{neurips_2026}

\usepackage[utf8]{inputenc} % allow utf-8 input

\usepackage[T1]{fontenc}    % use 8-bit T1 fonts
\usepackage{hyperref}       % hyperlinks
\hypersetup{hidelinks}
\usepackage{url}            % simple URL typesetting
\usepackage{booktabs}       % professional-quality tables
\usepackage{amsfonts}       % blackboard math symbols
\usepackage{nicefrac}       % compact symbols for 1/2, etc.
\usepackage{microtype}      % microtypography
\usepackage{xcolor}         % colors

\usepackage{tcolorbox}
\usepackage{pifont}
\usepackage{longtable}
\usepackage{makecell}
\usepackage{multirow} 
\usepackage{graphicx}
\usepackage{wrapfig}
\usepackage[title]{appendix}
\usepackage[accsupp]{axessibility}

\newcommand{\revised}[1]{#1}
\newcommand{\newrevised}[1]{#1}
\newcommand{\finalrev}[1]{#1}

\newcommand{\eg}[1]{{#1}}
\renewcommand{\acksection}{\section*{Acknowledgments}}

\title{Position: Next-Generation Game Engines Should Be Built on Interactive Generative Video}

\author{%
Jiwen Yu$^{1}$\thanks{Equal contribution.} \quad Yiran Qin$^{1*}$ \quad Haoxuan Che$^{2}$ \quad Quande Liu$^{3}$ \\
\textbf{Xintao Wang}$^{3}$\thanks{Corresponding authors.} \quad \textbf{Pengfei Wan}$^{3}$ \quad \textbf{Di Zhang}$^{3}$ \quad \textbf{Xihui Liu}$^{1\dagger}$ \\
$^{1}$The University of Hong Kong \\
$^{2}$The Hong Kong University of Science and Technology \\
$^{3}$Kuaishou Technology
}

\begin{document}

\maketitle

\begin{abstract}

Modern game development faces significant challenges in creativity and cost due to predetermined content in traditional game engines. Recent breakthroughs in video generation models, capable of synthesizing realistic and interactive virtual environments, present an opportunity to revolutionize game creation. \revised{This position paper argues that next-generation game engines should be built on Interactive Generative Video (IGV). We term such engines Generative Game Engines (GGE), with IGV as the perceptual and world-generation core under designer control.}
\revised{We argue why IGV is uniquely suited for this role, identify what is missing, and examine progress across six functional modules to show how each supports our position, culminating in a five-level maturity roadmap (L0--L4). Drawing on discussions with researchers, game developers, and community participants, we present nine alternative views and ethical considerations to stimulate debate.}
\end{abstract}
\section{Introduction}
% background #1: the challenge of game industry
Computer games face ever-growing market demand, yet the gaming industry faces three critical challenges: (1) current game engines rely on pre-made assets and fixed logic scripts, leading to predetermined content that players eventually exhaust; (2) existing engines cannot provide adaptive, personalized content tailored to individual players; (3) developing high-quality games, especially AAA titles, requires substantial resources and extensive development time. How to create high-quality games with unlimited personalized content while minimizing costs remains a fundamental challenge.

% background: Video Gen, World Model
During the past year, video generation models have made remarkable progress~\cite{sora,cogvideox,kling,veo2,seaweed,moviegen,runway,luma,vidu,wan,hunyuanvideo}, demonstrating capabilities in large-scale motion dynamics, semantic understanding, concept composition, 3D consistency with physical laws, and long-term temporal coherence. These advances show great potential for simulating real-world physics~\cite{sora,video-new-language,unisim}, suggesting these models could serve as world models for generating physically plausible videos.

% IGV definition
Building upon these advances, \revised{we identify and characterize an emerging technical direction we call} Interactive Generative Video (IGV), which extends video generation with interactive features incorporating four key characteristics: user control, memory of video context, understanding of physical rules, and causal reasoning intelligence. By combining these elements, IGV constructs an explorable and interactive virtual world, functioning similarly to a simulator.

% IGV can simulate existing games
The virtual worlds created by IGV naturally align with video games as they provide interactive environments where players explore and engage with dynamically generated content.
Recent works~\cite{genie,genie2,diamond,gamengine,gamegenx,matrix,playgen,gamefactory,oasis,wham,adaworld,mineworld,maag,worldmem} have demonstrated this potential by training action-conditioned video generation models on classic games like Atari~\cite{diamond}, DOOM~\cite{gamengine,playgen}, CS:GO~\cite{diamond}, Minecraft~\cite{gamefactory,oasis,mineworld,worldmem}, and Super Mario Bros~\cite{playgen}, creating interactive gaming experiences by iteratively generating video frames in response to user inputs.

% IGV can create new games
However, as pointed out by some works~\cite{gamefactory, matrix, genie2}, merely replicating existing games through IGV offers limited value over traditional game engines. The revolutionary potential of IGV lies in creating infinite entirely new games: everyone can become a game designer by providing design instructions to video generation models, which then generate limitless explorable virtual worlds, fundamentally transforming both game development and gaming experiences.

% propose our position
In conclusion, this position paper argues that \textbf{next-generation game engines should be built on Interactive Generative Video (IGV)}. \revised{We term such engines as Generative Game Engines (GGE), which dynamically generate both assets and logic through generative AI, with IGV serving as the perceptual and world-generation core under designer control.} Designers retain authority over style, rules, and the scope of generated variation. We argue that GGE can lower creation barriers and expand the range of authored experiences.

\revised{While the potential of video generation for interactive simulation is increasingly recognized, through extensive discussions with researchers, game developers, and community participants, we have found that substantial disagreement remains on key questions: whether IGV should serve as the \emph{core} technology for game engines, how it relates to traditional engines augmented with LLMs, whether 2D video generation or 3D scene generation should lead, and what the concrete path toward a fully functional generative game engine looks like. The diversity of alternative views we address in Sec.~\ref{sec:alternative} reflects the breadth of these open debates. Our contribution is a research agenda linking six capabilities, their unresolved dependencies, measurable maturity criteria, and competing views, rather than a new algorithm or a demonstration of a complete engine.}

\finalrev{Our core argument: video generation models offer generalizable world modeling, physics-aware generation, interactive control, and scalable data access, collectively enabling automated content creation that traditional engines require extensive manual effort to match (\emph{premise}; Sec.~\ref{sec:why}); though gaps remain (Sec.~\ref{sec:gap}), progress in memory, dynamics, reasoning, and gameplay motivates a research hypothesis that these gaps can be addressed (\emph{inference}; Sec.~\ref{sec:framework}); therefore IGV should serve as the foundational technology for next-generation game engines, and we chart a roadmap toward this goal (\emph{conclusion}; Sec.~\ref{sec:levels}).}

\revised{This paper is organized as follows. Sec.~\ref{sec:preliminary} introduces preliminaries. Sec.~\ref{sec:why} argues why IGV is uniquely suited as the core technology for GGE. Sec.~\ref{sec:gap} identifies what is still missing from IGV to a full game engine. Sec.~\ref{sec:framework} examines technical progress across six modules to argue why each supports the feasibility of our position. Sec.~\ref{sec:levels} charts a roadmap toward fully generative game engines. Finally, Secs.~\ref{sec:alternative}--\ref{sec:conclusion} address alternative perspectives, ethical issues, and concluding remarks.}

\section{Preliminaries}
\label{sec:preliminary}
A more detailed preliminary section can be found in Appendix~\ref{appendix:preliminary}.

\textbf{Video Generation Models.} Video generation models have achieved significant breakthroughs with the rise of diffusion models~\cite{ddpm, sde, score,flow,rectified}, which have become the mainstream approach due to their superior generation quality~\cite{sora,cogvideox,kling,veo2,seaweed,moviegen,runway,luma,vidu,wan,hunyuanvideo}. 
The field has also made substantial progress in conditional video generation~\cite{guo2025sparsectrl, ni2023conditional, dynamicrafter}, particularly in camera control~\cite{direct-a-video, motionctrl, cameractrl, 3dtrajmaster, recammaster}, where methods like MotionCtrl~\cite{motionctrl} and CameraCtrl~\cite{cameractrl} enable precise manipulation of camera movements. For autoregressive video generation, which is crucial for creating variable-length or infinite video sequences, two representative approaches have emerged: GPT-like next-token prediction methods~\cite{videopoet,nova,emu3} and Diffusion Forcing~\cite{cdf,dfot}.

\textbf{AI-driven Game Applications.} AI technologies have demonstrated diverse applications in game creation. In game video generation, recent works leveraging diffusion models~\cite{diamond,gamengine,oasis} have achieved high-quality results, with open-domain methods~\cite{genie2,matrix,gamefactory} even enabling the creation of novel game content. AI-powered design assistants have enhanced the game development process by automating design completion~\cite{tanagra} and generating multiple design suggestions~\cite{sketchbook,susketch}, thereby streamlining development and fostering creativity. Furthermore, intelligent game agents have evolved from traditional reinforcement learning approaches~\cite{gsb,clip4mc} to more sophisticated LLM-based methods~\cite{voyager,deps}, significantly improving performance in long-horizon tasks.
\section{Why Interactive Generative Video for Game Engines?}
\label{sec:why}
Computer games face ever-growing market demand, yet developing high-quality games requires substantial resources. Traditional engines like \textit{Unreal} and \textit{Unity} rely on pre-made assets and fixed logic scripts, limiting content to predefined scenes that players will eventually exhaust. Creating innovative games at scale while minimizing costs remains a critical challenge.

\revised{We frame the} Generative Game Engine (GGE) as a next-generation solution that dynamically generates both assets and logic, aiming to (1) reduce asset-authoring effort, (2) lower creation barriers, and (3) support open-ended, dynamically generated worlds. These potential benefits do not establish lower total costs or better player experiences in every setting.

Building upon recent advances in video generation, \revised{we identify} Interactive Generative Video (IGV) as a promising foundation for GGE. As illustrated in Fig.~\ref{fig:framework} (a), IGV is viewed from a technical angle (video generation with interactive control), while GGE is viewed from an application angle (a game engine utilizing generative AI). IGV offers four key advantages: (1) generalizable generative capabilities, (2) physics-aware world modeling, (3) user-controlled interactive generation, and (4) leveraging vast video data for training. We elaborate on these below.

\revised{By ``core,'' we mean the foundational substrate for world representation and perceptual rendering: learned interactive video generation is the primary way the engine produces the dynamic visual world. This is an architectural claim, beyond using AI to generate assets for a conventional renderer. It does not require video to store every state variable or replace every simulator. Designers steer generation through prompts, control signals, explicit rules, and constraints; LLMs and other components can supply high-level logic. IGV-as-core is therefore compatible with hybrid state (Sec.~\ref{subsec:memory}) and with authored games that preserve a director's style and intent.}

% \subsection{High-quality Content Generation from Videos}
% Video generation models demonstrate exceptional capabilities in creating high-quality visual content. These capabilities stem from their training on vast collections of real-world video data, which enables them to learn rich representations of visual elements. The models exhibit remarkable compositional generation abilities by deeply understanding and creatively combining elements from their training data to synthesize novel, coherent, and unprecedented scenes, offering unique advantages in content creation. These capabilities directly benefit game development by reducing content creation costs while meeting the high quality standards of modern game production. 
% As demonstrated in Figure~\ref{fig:generation}, state-of-the-art commercial video generation models already showcase impressive generation capabilities, validating their potential for revolutionizing game content creation.

% \subsection{Generalizable Content Generation for Unlimited Novel Games}
\subsection{Generalizable Generation for Unlimited Games}
%  \begin{figure}[!tbp]
%   \centering
%   % \vspace{-0.1cm}
%   \includegraphics[width=1\linewidth]{figure/gf_demo.jpg}
%   \vspace{-0.7cm}
%   \caption{Demonstration of GameFactory~\cite{gamefactory}'s ability to generalize action control capabilities learned from Minecraft data to open-domain scenarios. Examples from its homepage showcase various generalized environments where the learned control mechanisms have been successfully applied.}
%   % \vspace{-0.5cm}
% \label{fig:generalization} 
% \end{figure}

\begin{wrapfigure}[13]{r}{0.4\textwidth}
\vspace{-0.5cm}
\includegraphics[width=0.4\textwidth]{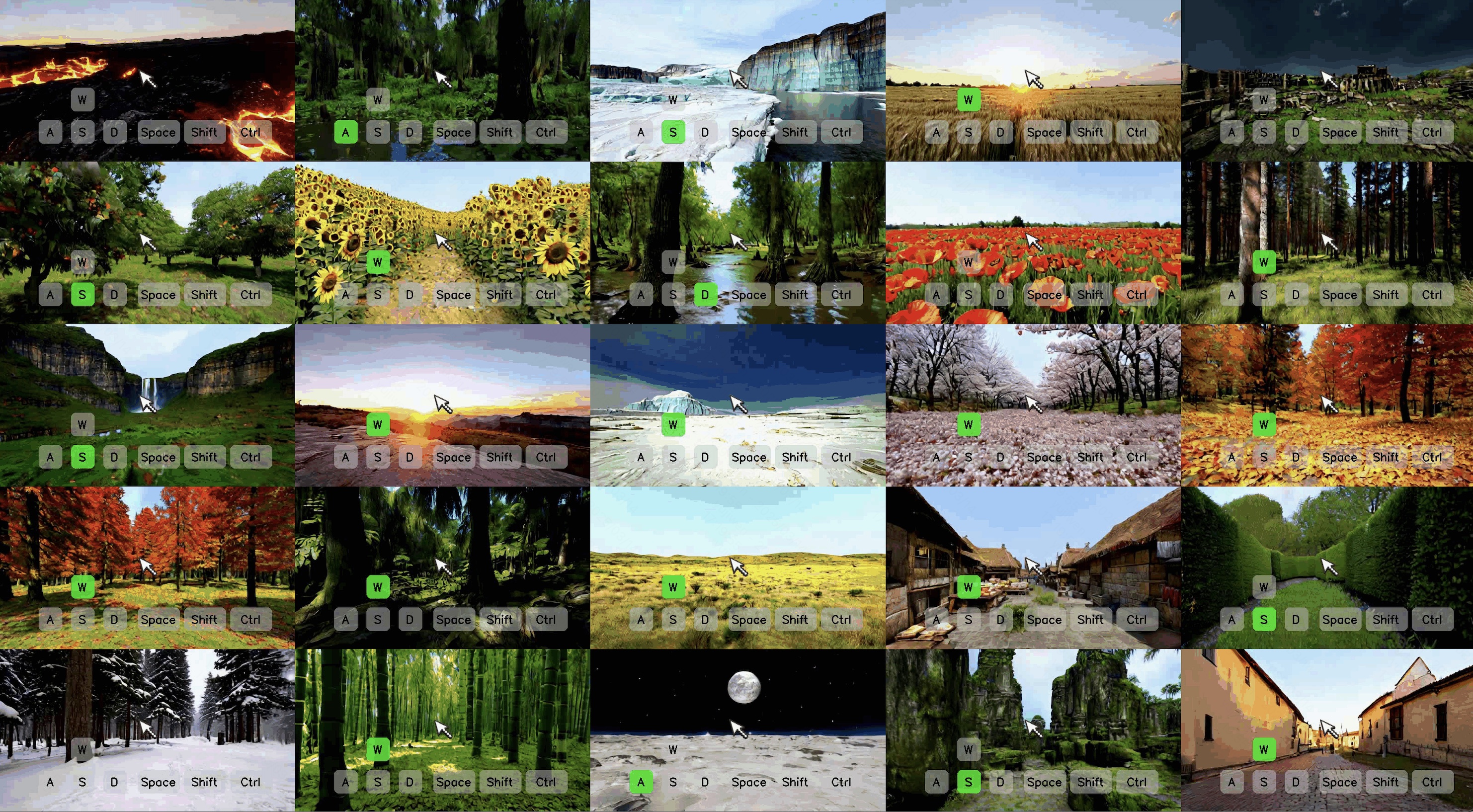}
% \vspace{-0.5cm}
\caption{Demonstration of GameFactory~\cite{gamefactory}'s ability to generalize action control abilities learned from Minecraft data to open-domain scenarios. }
% \vspace{3cm}
\label{fig:generalization}
\end{wrapfigure}
Video generation models excel at creating not just high-quality content, but novel and diverse game content. Pre-trained on vast real-world video collections, these models develop comprehensive understanding of visual elements and relationships. Their novelty manifests in two aspects: (1) generalization ability to transfer skills to unprecedented scenarios, as shown by GameFactory~\cite{gamefactory} generating action-controllable videos in open-domain settings (Fig.~\ref{fig:generalization}), and (2) compositional creativity to combine learned elements innovatively, demonstrated by Sora~\cite{sora}'s "origami undersea" scenes\footnote{\url{https://www.youtube.com/watch?v=KGcLSTFEgSk}}. This compositional capability has become a key research focus, with dedicated evaluation benchmarks~\cite{t2icomp, t2vcomp} measuring such abilities.

\subsection{Physics-aware World Modeling}
%  \begin{figure}[!tbp]
%   \centering
%   % \vspace{-0.1cm}
%   \includegraphics[width=1\linewidth]{figure/physics.pdf}
%   \vspace{-0.7cm}
%   \caption{Physics-aware generation capabilities of video models. Top: Examples from Cosmos~\cite{cosmos} demonstrating physical understanding in diverse scenarios including robotics, autonomous driving, manufacturing, and home environments. Bottom: Human motion examples generated by Kling.}
%   % \vspace{-0.5cm}
% \label{fig:physics} 
% \end{figure}

\begin{wrapfigure}[14]{r}{0.4\textwidth}
\vspace{-0.5cm}
\includegraphics[width=0.4\textwidth]{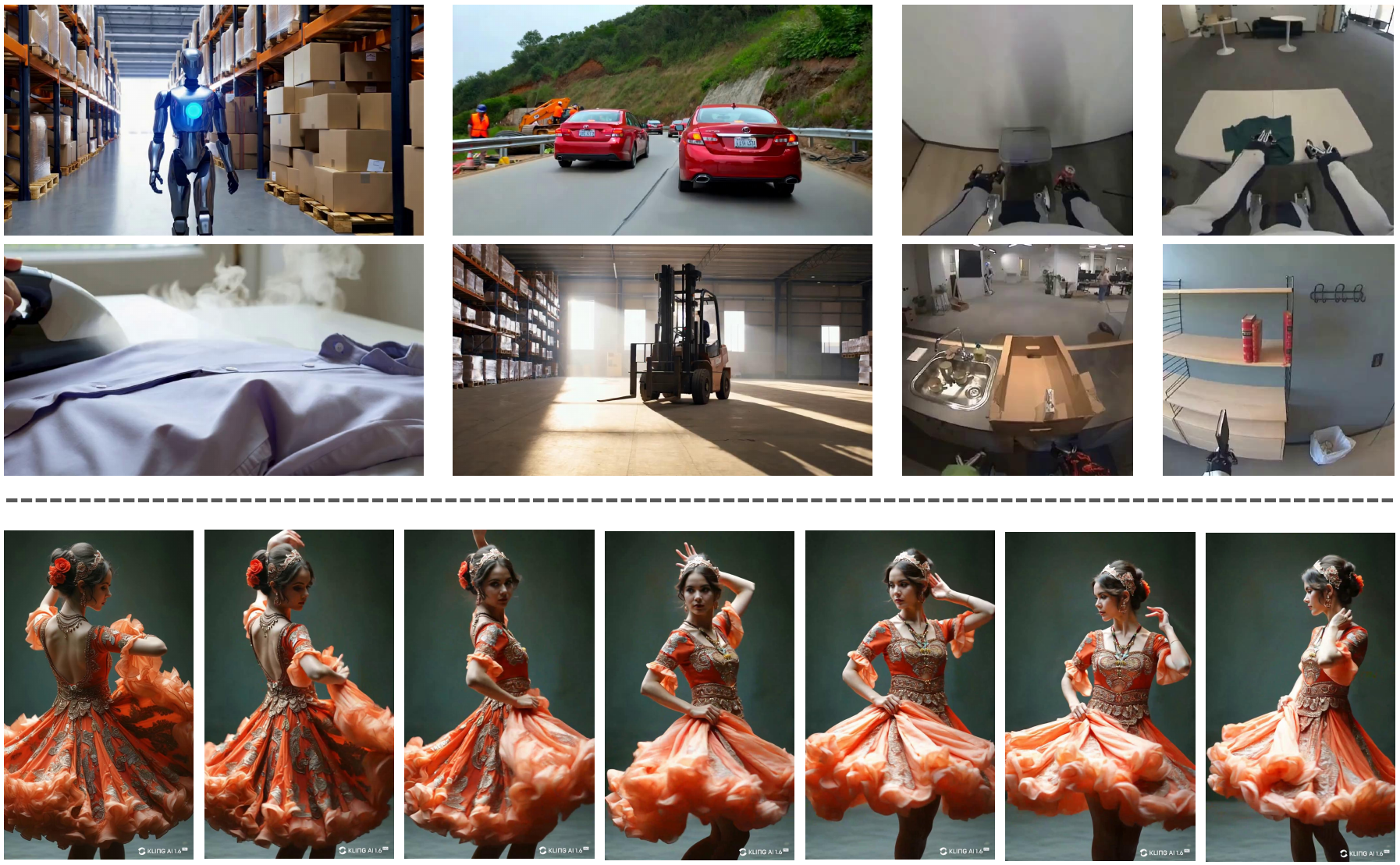}
\vspace{-0.5cm}
\caption{Physics-aware video generation. Top: Cosmos~\cite{cosmos} examples across robotics, driving, and home scenarios. Bottom: Human motion by Kling~\cite{kling}.}
% \vspace{3cm}
\label{fig:physics}
\end{wrapfigure}
Video generation models demonstrate remarkable potential in understanding real-world physical knowledge~\cite{sora,video-new-language,unisim}. During training, these models naturally learn implicit physical priors encompassing gravity, elasticity, collisions, and complex motion patterns. While traditional game engines rely on predefined physical formulas or motion capture, IGV leverages learned priors to directly generate physically plausible content (Fig.~\ref{fig:physics}), simplifying engine design and lowering developer expertise requirements.
\revised{We acknowledge that recent benchmarks~\cite{how-far-world-model,Physics-IQ} have revealed significant limitations in current video models' physical understanding, particularly for out-of-distribution scenarios and formula-based rigid body mechanics. However, IGV's physics advantage lies not in replacing explicit physical solvers, but in covering \emph{formula-hard} phenomena (fluid deformation, cloth wrinkling, soft-body dynamics) and \emph{semantic} physics that are difficult or impossible to express with traditional formulas. We discuss this complementary relationship further in Alternative View \#7 (Sec.~\ref{sec:alternative}). Physical plausibility is relative to the intended world: stylized motion, impossible gravity, or other deliberately non-physical rules should be preserved through designer constraints, rather than overwritten by real-world priors.}

\begin{figure}[t]
\centering
\includegraphics[width=\linewidth]{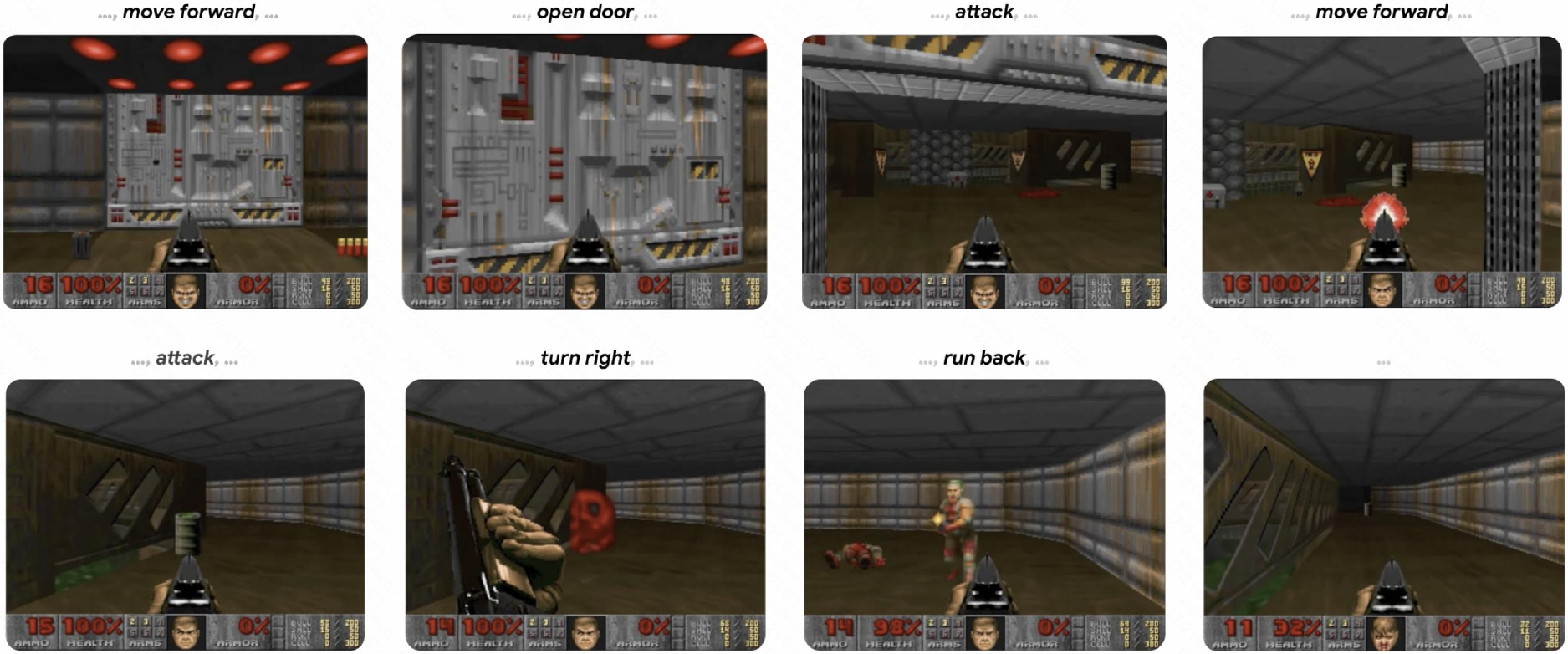}
\caption{Interactive gameplay in GameNGen~\cite{gamengine}. Top row, left to right: move forward, open door, attack, move forward. Bottom row: attack, turn right, run back, and the subsequent frame.}
\label{fig:gamengine}
\end{figure}

\subsection{Interactive Generation with User Control}
Precise control of visual generation models has made significant progress~\cite{t2iadapter,controlnet,control-a-video,controlnext}, supporting camera viewpoint adjustment~\cite{motionctrl,cameractrl,3dtrajmaster,recammaster}, character movement~\cite{animate-anyone}, and increasingly diverse control signals~\cite{fulldit,drag-a-video} (Fig.~\ref{fig:gamengine}).
\revised{Notably, IGV's control advantage is strongest for complex interactions (e.g., picking up objects, operating tools) that traditional engines can only support by pre-building dedicated animations for each action; IGV achieves these by simply generating plausible video. We discuss this further in Alternative View \#7 (Sec.~\ref{sec:alternative}).}

\subsection{Video Data Accessibility Enables Scaling}

Video data offers unique advantages through its accessibility and unified representation format~\cite{video-new-language}. Unlike traditional engines that require heterogeneous assets (3D models, textures, animations) with substantial manual effort, videos are widely available and continuously growing across internet platforms. Videos naturally capture diverse real-world phenomena, enabling models to learn comprehensive world knowledge at scale. Using video as a unified representation avoids the complexity of managing different asset formats. \revised{We note that raw visual corpora are abundant and provide strong priors, while action-conditioned video pairs essential for interactive control are currently scarcer and remain an active research area.}

\revised{\section{What Is Still Missing from IGV to a Game Engine?}
\label{sec:gap}

Sec.~\ref{sec:why} established that IGV offers compelling advantages as a foundation for next-generation game engines. However, a video generation model alone is not yet a game engine. We must identify what additional capabilities are needed.

A game engine must support \emph{sustained interaction}: players expect to play for hours, not seconds, demanding streaming real-time generation far beyond current clip-level synthesis. It must maintain \emph{persistent memory}: a building constructed ten minutes ago should still be there when the player returns. It must respect \emph{world rules}: objects should fall, collide, and break in predictable ways, tunable by designers rather than merely emergent from training data. It must exhibit \emph{higher-order intelligence}: NPCs should reason about causality, and the world should be capable of self-evolution. \newrevised{It must support \emph{multiplayer consistency}: simultaneous inputs from multiple players must be reconciled into a single coherent world state under strict latency constraints.} Finally, games require explicit \emph{gameplay rules} (objectives, rewards, progression) imposed by designers on top of the virtual world.

These requirements naturally lead to a modular decomposition. We identify six functional modules for a complete Generative Game Engine: \textbf{Generation} (the base video synthesis capability), \textbf{Control} (multi-modal interactive input), \textbf{Memory} (static and dynamic state persistence), \textbf{Dynamics} (physical rule modeling and tuning), \textbf{Intelligence} (causal reasoning and self-evolution), and \textbf{GamePlay} (designer-imposed game rules). The first five modules, centered on IGV, together create an interactive virtual world; the GamePlay module transforms that world into a playable game.

We emphasize that this decomposition is a \emph{conceptual} tool for identifying research bottlenecks and providing entry points for different AI sub-communities (RL, 3D vision, NLP, etc.), not a prescription for the final system architecture. The ultimate implementation may well converge toward a unified, video-native multimodal world model that internalizes these capabilities end-to-end. In the next section, we examine each module's current progress and remaining challenges to explain why these gaps merit research toward IGV-centered engines, while distinguishing partial progress from a demonstrated solution.}

\section{What Progress Supports the Feasibility of IGV-Centered Engines?}
\label{sec:framework}
\begin{figure}[t]
  \centering
%   \vspace{-0.1cm}
  \includegraphics[width=1\linewidth]{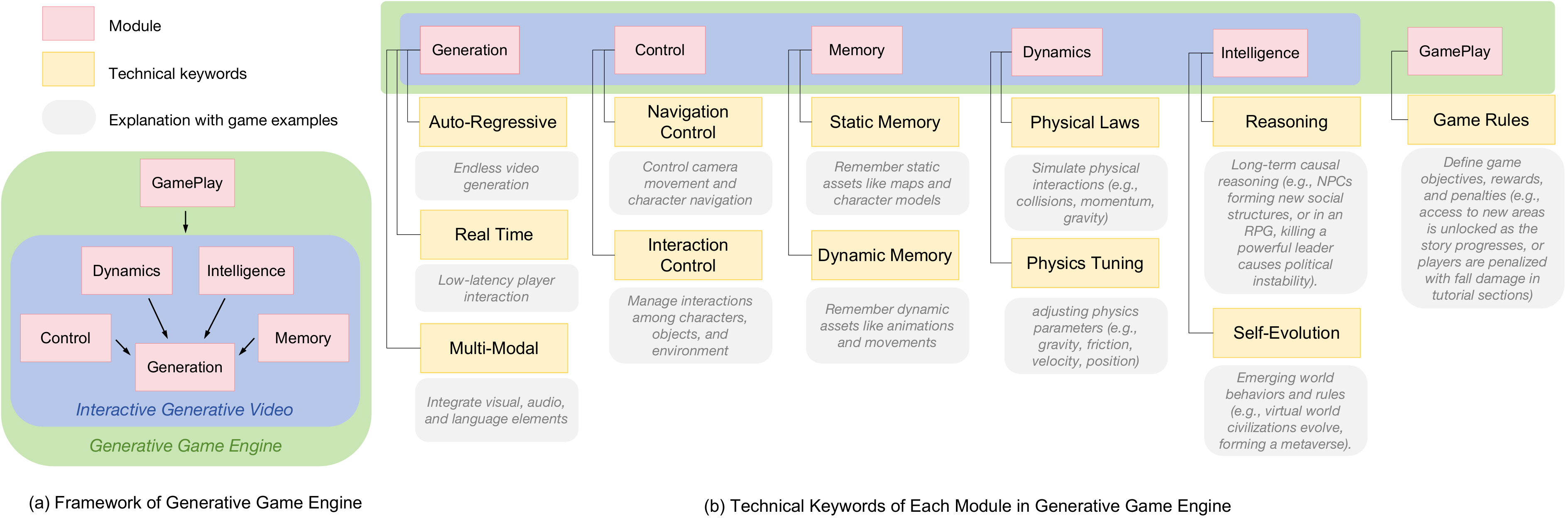}
  \vspace{-0.6cm}
  \caption{Proposed framework of Generative Game Engine (GGE).
  (a) Architecture and interactions between modules of GGE. (b) Technical keywords and their explanation of each module. 
  % The game examples shown in gray boxes demonstrate typical applications of each module's capabilities, with detailed analysis provided in the main text.
  }
  \vspace{-0.5cm}
\label{fig:framework} 
\end{figure}

% \begin{wrapfigure}[17]{r}{0.4\textwidth}
% \vspace{-0.5cm}
% \includegraphics[width=0.4\textwidth]{figure/framework.pdf}
% \vspace{-0.5cm}
% \caption{Proposed framework of Generative Game Engine (GGE).
%   (a) Framework of GGE shows the architecture and interactions between modules. (b) Technical keywords of each module. 
%   The game examples shown in gray boxes demonstrate typical applications of each module's capabilities, with detailed analysis provided in the main text.}
% % \vspace{3cm}
% \label{fig:framework}
% \end{wrapfigure}

Fig.~\ref{fig:framework} illustrates the architecture of our proposed GGE framework. Fig.~\ref{fig:framework}~(a) shows the relationships between the six modules identified in Sec.~\ref{sec:gap}, and Fig.~\ref{fig:framework}~(b) presents their key technical components. Below we examine each module's role, evidence, and remaining gaps. Progress on individual capabilities motivates our position; jointly achieving latency, control, memory, and consistency remains unproven. Named games illustrate desired functions, not capabilities already reproduced by current IGV systems.

\subsection{Generation}
\label{subsec:generation}
\textbf{\ding{113} Concept.} 
The Generation Module handles video generation, the fundamental functionality of IGV. Beyond ensuring visual quality and motion coherence, it encompasses three crucial functionalities:
(1) \textbf{Streaming Generation} enables continuous video synthesis with frame-level control frequency. \eg{This supports endless procedural worlds in \textit{No Man's Sky}, real-time weather cycles in \textit{Red Dead Redemption 2}, and instant response in rhythm games like \textit{Beat Saber}}.
(2) \textbf{Real-time Processing} facilitates low-latency interaction. \eg{This is essential in competitive games like \textit{Counter-Strike} and \textit{League of Legends} where instant visual feedback is crucial}.
(3) \textbf{Multi-modal Generation} complements video with other modalities. \eg{This includes dynamic music in \textit{Journey}, positional audio in \textit{PUBG}, and real-time dialogue subtitles in \textit{Mass Effect}}.

\textbf{\ding{113} Technical Approaches and Future Directions.} 
% The implementation and future development of the Generation Module centers around the three key functionalities mentioned above:

(1) \textbf{Streaming Generation}:

Diffusion-based methods~\cite{sora} excel at generating high-quality visual content. Streaming generation can be achieved by applying different noise levels across frames, allowing later frames (higher noise) to condition on previous frames (lower noise); representative methods like Diffusion Forcing~\cite{cdf,dfot,causvid,ar-diffusion,causalfusion} have been widely used in game video generation~\cite{matrix,gamengine,oasis,gamefactory}. Next token prediction~\cite{videopoet,emu3} lags in visual quality but offers LLM integration for causal reasoning~\cite{zhou2024transfusion,xie2024show}. Hybrid approaches combining both~\cite{mar, nova} aim to maintain quality while modeling frame causality, though they remain in early stages.

\revised{A critical challenge is \textbf{long-horizon drift}: error accumulation over extended autoregressive rollouts leads to visual degradation and semantic inconsistency. Recent works~\cite{self-forcing,longlive,rolling-forcing,svi} have begun addressing this, but mitigating drift at game-relevant scales remains an open problem.}

(2) \textbf{Real-time Generation}: 

Recent advances include model distillation~\cite{bk-sdm}, diffusion step reduction~\cite{videolcm}, high-compression VAEs~\cite{dcae}, and causal architectures like CausVid~\cite{causvid} and Cosmos~\cite{cosmos}. \revised{CausVid reports approximately 9.4 fps on a single GPU under its evaluation conditions. This is evidence of efficiency progress, not a direct estimate of the work needed for 60 fps at 1080p with reliable control and long-horizon consistency. Slower-paced narrative and turn-based experiences may tolerate lower frame rates; offline level pre-construction is another possible use.} Frame rate, input-to-display latency, resolution, and deployment cost must be evaluated together.

(3) \textbf{Multi-modal Generation}: 

One approach is to develop unified multimodal models that support understanding and generation across text, vision, audio, and other modalities~\cite{zhou2024transfusion,xie2024show,videopoet,gato}. An alternative is to first develop specialized models per modality~\cite{pengi,depthanything,motiongpt,lvm} and then compose them into a pipeline where, for example, language models produce generation instructions, video models synthesize the visual output, and audio models generate corresponding sound.

\textbf{\ding{113} Why This Supports Our Position.}
Distillation and causal generation provide concrete avenues for improving throughput, supporting research on real-time generative engines. They do not establish a known scaling trajectory or show that latency is merely an engineering issue. Quality, controllability, and long-horizon stability may introduce additional limits; our tractability claim is a reasoned research hypothesis, not a solved result.

\vspace{-0.5cm}

\subsection{Control}
%  \begin{figure}[!tbp]
%   \centering
%   \vspace{0.1cm}
%   \includegraphics[width=0.8\linewidth]{figure/control.pdf}
%   \vspace{-0.3cm}
%   % \caption{\textbf{Control modules.} The left figure illustrates the player's navigation control during exploration, while the right figure demonstrates the player's interaction control with the environment when striking a rock.
%   \caption{The Control module of IGV manages player control through two aspects: Navigation Control and Interaction Control.}
%   % \vspace{-0.5cm}
% \label{fig:control} 
% \end{figure}

\begin{wrapfigure}[17]{r}{0.4\textwidth}
\vspace{-0.5cm}
\includegraphics[width=0.4\textwidth]{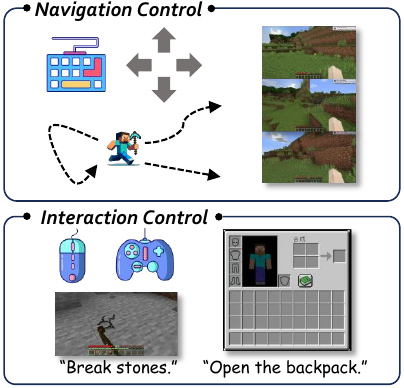}
\vspace{-0.5cm}
\caption{The Control module manages player control through two aspects: navigation and interaction control.}
% \vspace{3cm}
\label{fig:control}
\end{wrapfigure}
\textbf{\ding{113} Concept.} 
As shown in Fig.~\ref{fig:control}, the control module manages user control of the virtual world through two aspects:
(1) \textbf{navigation control} enables camera and character movement \eg{(e.g., WASD for movement, mouse for camera rotation, and space bar for jumping in open-world games)}.
(2) \textbf{interaction control} allows players to manipulate objects within the virtual environment \eg{(e.g., clicking to place or rotate buildings, keyboard shortcuts to access inventory in construction games)}.

\textbf{\ding{113} Technical Approaches and Future Directions.} 

Common approaches include Cross Attention~\cite{gamefactory,gamengine,matrix} and Adaptors~\cite{gamegenx,motionctrl} for fusing control signals with video features. While control is easily mastered in fixed scenes, open-domain generalization with limited annotations remains challenging~\cite{gamefactory,matrix,genie2}.

Interaction control requires understanding how interactions change the environment (related to Sec.~\ref{subsec:dynamics}), motivating large-scale data collection~\cite{gamefactory,gamegenx}. A promising direction is natural control signals (e.g., gesture recognition, brain-computer interfaces) aligning with users' gaming intuitions.

\revised{\textbf{Multi-agent Control.}
Multiplayer support is fundamentally a multi-agent control problem: simultaneous inputs from multiple players must be reconciled into a single consistent world state. Recent works~\cite{solaris,multigen,multiworld} have begun exploring multiplayer video world models with cross-view coherence and shared memory. Key challenges include the tension between generative stochasticity and deterministic state synchronization, and integration with traditional netcode infrastructure (rollback, server authority, lag compensation). }

\textbf{\ding{113} Why This Supports Our Position.}
Control capabilities learned from a single game (e.g., Minecraft) generalizing to open-domain scenarios demonstrate that IGV's interactive control is not game-specific but transferable, unlike traditional engines whose control logic is hard-coded per game. Recent multi-agent control research also offers approaches to multiplayer consistency, although robust synchronization under concurrent inputs remains unresolved.

\subsection{Memory}
\label{subsec:memory}
%  \begin{figure}[t]
%   \centering
%   \vspace{0.1cm}
%   \includegraphics[width=0.8\linewidth]{figure/Memory.pdf}
%   \vspace{-0.3cm}
%   \caption{The Memory module of IGV consists of two components: static memory and dynamic memory.}
%   % \vspace{-0.5cm}
% \label{fig:Memory} 
% \end{figure}

\begin{wrapfigure}[14]{r}{0.4\textwidth}
\vspace{-0.4cm}
\includegraphics[width=0.4\textwidth]{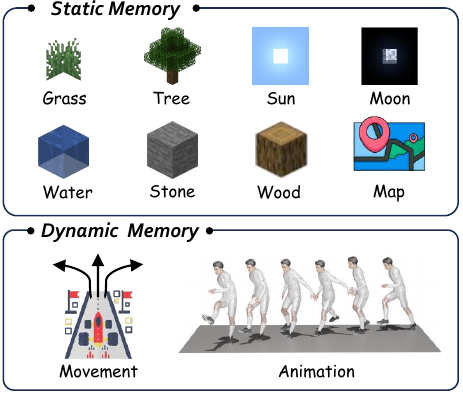}
\vspace{-0.5cm}
\caption{The Memory module consists of static and dynamic memory.}
% \vspace{3cm}
\label{fig:Memory}
\end{wrapfigure}
\textbf{\ding{113} Concept.} Conventional video generation models rely solely on attention mechanisms, struggling to maintain visual consistency in long-duration or large-motion scenarios. As shown in Fig.~\ref{fig:Memory}, the Memory module addresses this through two aspects: (1) \textbf{static memory} encompasses scene- and object-level memory (game maps, buildings, character models)\eg{; inconsistency in player-built structures (\textit{Minecraft}, \textit{SimCity}) would severely impact experience}. (2) \textbf{dynamic memory} handles short-term motion and behavior patterns (character animations, vehicle trajectories, weather transitions)\eg{, crucial in fighting games where attack animations must remain coherent or rhythm games requiring smooth transitions}.
% However, it's important to note that long-term memory requiring complex logical reasoning falls beyond the scope of this module. For instance, when an NPC in a game is assigned a task by the player, completes it over an extended period, and returns with results, this process requires more than simple memory capabilities. Such advanced cognitive functions are better addressed in the GamePlay Module, which we will discuss later.

\textbf{\ding{113} Technical Approaches and Future Directions.} 

Current methods rely on attention-based memory via cross-attention between historical and predicted frames~\cite{gamengine,oasis,worldmem}, but face limitations in precision and window size.
Dedicated memory structures offer a promising alternative, implemented as implicit high-dimensional features~\cite{gamegan} or explicit 3D representations~\cite{pe,pgm,wonderjourney,viewcrafter,see3d,gen3c} for consistent preservation of static elements.

\revised{We favor a \textbf{hybrid world state}: frame history preserves visual detail, latent memory offers compact long-context information, explicit 3D/spatial structures impose geometric constraints, and symbolic variables record rules such as health, inventory, and progression. Developer interfaces should support querying, editing, inspecting, and saving/restoring this state, directly or through higher-level tools and agents. Maintaining agreement among these representations after edits, and restoring state without drift, remain open systems challenges.}
% Dynamic memory, which needs to record animations, actions and trajectories, requires dynamic video datasets, especially those containing significant motions like human movements. Collecting and annotating such high-quality dynamic data remains challenging but crucial for advancing this research direction.

\textbf{\ding{113} Why This Supports Our Position.}
Dedicated memory structures demonstrate that IGV need not rely solely on finite attention windows. Explicit spatial memory offers a route to persistent visited regions while retaining generative flexibility for unseen areas; comparable reliability to conventional scene graphs is still a target.
\newrevised{KV cache reuse, spatial memory, and distillation may reduce redundant computation, but their net savings must be measured against optimized traditional rendering at comparable quality.}

\subsection{Dynamics}
\label{subsec:dynamics}
%  \begin{figure}[t]
%   \centering
%   \vspace{0.1cm}
%   \includegraphics[width=0.8\linewidth]{figure/Dynamics.pdf}
%   % \vspace{-0.7cm}
%   \caption{The Dynamics module of IGV focuses on two
% aspects: physical laws and physics tuning.}
%   % \vspace{-0.5cm}
% \label{fig:Dynamics} 
% \end{figure}

\begin{wrapfigure}[15]{r}{0.4\textwidth}
\vspace{-0.5cm}
\includegraphics[width=0.4\textwidth]{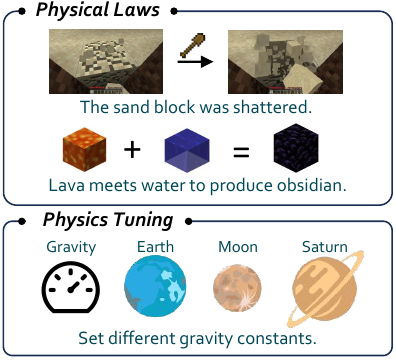}
\vspace{-0.5cm}
\caption{The Dynamics module focuses on physical laws and physics tuning.}
% \vspace{3cm}
\label{fig:Dynamics}
\end{wrapfigure}

\textbf{\ding{113} Concept.} 
As shown in Fig.~\ref{fig:Dynamics}, the Dynamics Module focuses on two aspects:
(1) \textbf{Physical Laws}: generating videos compliant with fundamental physics (gravity, collision, acceleration)\eg{, driving core mechanics in simulators like \textit{Forza Motorsport} and puzzles like \textit{Portal}}.
(2) \textbf{Physics Tuning}: control over physical parameters rather than replicating real-world physics (e.g., adjusting gravity, friction, time)\eg{, as in \textit{Braid} and \textit{Superhot} which rely on manipulated physics for unique gameplay}.

\textbf{\ding{113} Technical Approaches and Future Directions.} 
% The implementation and future development can be analyzed from two main aspects:

A data-driven approach learns physical laws from large-scale video data~\cite{sora,cosmos}, though requiring diverse physical video data~\cite{zhao2025synthetic}. An alternative uses video generation models as renderers on top of physics simulators~\cite{physgen,diffusion-as-shader}, using explicit physical constraints, while visual compliance with the simulated state still needs verification. Benchmarks for physical accuracy remain crucial~\cite{worldsimbench,worldmodelbench}. Physics tuning, often overlooked, is a key open direction; synthetic data with annotated parameters is promising.

\textbf{\ding{113} Why This Supports Our Position.}
The complementary architecture, where explicit physics solvers handle formula-based mechanics and IGV handles formula-hard phenomena (fluids, cloth, soft-body), means that GGE does not need to reinvent physics from scratch; it inherits decades of physics engine research as a pluggable control signal while adding capabilities that traditional engines struggle to provide. This hybrid design narrows the learning problem, although reliable agreement between generated observations and simulated state remains a research challenge.

\subsection{Intelligence}

%  \begin{figure}[!tbp]
%   \centering
%   \vspace{0.1cm}
%   \includegraphics[width=0.8\linewidth]{figure/Intelligence.pdf}
%   \vspace{-0.3cm}
%   \caption{The Intelligence module of IGV implements two aspects: causal reasoning and self-evolution.}
%   % \vspace{-0.5cm}
% \label{fig:Intelligence} 
% \end{figure}

\begin{wrapfigure}[15]{r}{0.4\textwidth}
\vspace{-0.5cm}
\includegraphics[width=0.4\textwidth]{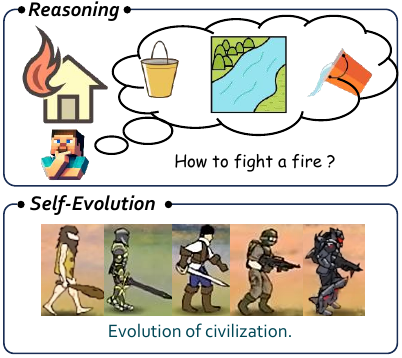}
\vspace{-0.5cm}
\caption{The Intelligence module implements reasoning and self-evolution.}
% \vspace{3cm}
\label{fig:Intelligence}
\end{wrapfigure}

\textbf{\ding{113} Concept.}
As shown in Fig.~\ref{fig:Intelligence}, the Intelligence module implements two aspects:
(1) \textbf{Reasoning}: long-term causal inference based on initial conditions. \eg{For example, predicting how a kingdom's economy evolves over centuries (\textit{Crusader Kings}) or simulating wildlife migration when environments change (\textit{Planet Zoo}).}
(2) \textbf{Self-Evolution}: virtual worlds continuously develop new knowledge, rules, and behaviors through emergent properties\eg{, e.g., civilizations forming cultures, ecosystems developing new species, cities growing organically}.

\textbf{\ding{113} Technical Approaches and Future Directions.}

Implementing reasoning requires causal autoregressive generation (Sec.~\ref{subsec:generation}) and large-scale long-context training~\cite{far}, or alternatively leveraging multimodal LLMs for causal reasoning alongside video generation~\cite{zhou2024transfusion,xie2024show}. If these capabilities are successfully combined, self-evolution may emerge, enabling autonomous virtual worlds.

\textbf{\ding{113} Why This Supports Our Position.}
Combining autoregressive video generation with LLM-based reasoning offers a path toward novel, contextually appropriate consequences beyond explicitly authored branches. This motivates the Intelligence module; it does not yet demonstrate reliable long-horizon causal reasoning. Self-evolution must also respect designer-defined rules and identity (Sec.~\ref{sec:levels}).

\subsection{Gameplay}

\textbf{\ding{113} Concept.}
The GamePlay Module builds upon IGV by implementing external \textbf{Game Rules}, which are designer-imposed rules such as game objectives, rewards, penalties, and constraints that shape the virtual world's gameplay experience. \eg{These include scoring systems in \textit{Tetris}, health and damage systems in \textit{Dark Souls}, time limits in \textit{Mario}, and competitive ranking systems in \textit{League of Legends}.}

\textbf{\ding{113} Technical Approaches and Future Directions.} 

Exact rules need not be inferred from raw video: video priors supply appearance and interaction patterns, while designers specify scoring, health, progression, and constraints. The granularity of learnable mechanics depends on the data and controls; no universal level is established. The GamePlay module may combine explicit rule logic with agent systems empowered by LLMs~\cite{gpt4,llama} or multimodal models~\cite{llava}, enabling level design, difficulty scaling, and NPC development. While communicative multi-agent frameworks~\cite{camel,autogen} show promise, their integration with generative game environments remains an open challenge. \revised{Important open challenges include dynamic adaptive game rules, automated playtesting, designer tooling beyond prompting, and principled reward/difficulty scheduling. GamePlay remains the least mature module in GGE; a full treatment of game design (core loop, pacing, retention) is beyond our engine-technology focus, but we flag it as a critical gap for future work.}

\textbf{\ding{113} Why This Supports Our Position.}
The GamePlay module illustrates why the critique ``video generation is not a game'' confuses the foundation with the full system. Just as Unreal separates rendering from gameplay logic (Blueprint sits atop the rendering pipeline), GGE separates world generation (IGV) from rule imposition (GamePlay). Agent-based rule specification offers one possible interface, but faithfully enforcing rules in the generated world remains an open requirement. \finalrev{Without this separation, a generative world would lack the structured goals that distinguish a game from a sandbox.}

\section{A Roadmap Toward Fully Generative Game Engines}
\label{sec:levels}
We propose a five-level maturity model (L0--L4) to evaluate GGE and guide their future development. This framework helps assess current technologies and identify key research directions in GGE. 
Below we detail each level. The overview table is presented in Appendix~\ref{appendix:levels}, and we provide concrete observable criteria for each level in Appendix~\ref{appendix:criteria} to make this roadmap measurable and actionable.
\newrevised{We offer L2 in roughly 3--5 years and L4 in roughly 10--15 years from the time of writing (2026) as preliminary, conditional forecasts, not guarantees. These agenda-setting estimates depend on unresolved advances in latency, memory, consistency, and control, and should be revised as evidence accumulates.}

\textbf{Level 0: No AI-Assisted Assets Generation.}
All assets and rules are manually crafted and fixed\eg{ (\textit{Super Mario}, \textit{Tetris})}. Precise control, but heavy resources and no dynamic content.

% \textbf{Level 1: Offline AI-Assisted Assets.}
% The first step towards generative capabilities involves using AI tools to create game assets during the development phase. While the integration of these assets still requires manual effort, this approach significantly reduces the workload of content creation. For instance, in \textit{Cyberpunk 2077}, developers can utilize image generation models like Stable Diffusion~\cite{sd} to create diverse textures for neon billboards, trash piles, and urban details throughout Night City. These AI-generated assets are carefully curated and integrated by developers before the game's release, accelerating the game development process.

% \textbf{Level 2: Semi-real-time Clip Generation.}
% At this level, the game engine can generate short segments in response to player actions. Unlike Level 1, this generation occurs during gameplay, but is limited to discrete, self-contained clips. When a player triggers specific events, such as destroying a bridge in an open-world game, the engine generates a 10-second explosion animation sequence. This real-time generation adds variability to common game events.

\textbf{Level 1: AI-Assisted Assets Generation.}
AI assists asset and logic creation\eg{ (e.g., Stable Diffusion~\cite{sd} for textures, dynamic NPC dialogues in \textit{AI Dungeon})}, but the framework remains pre-designed with significant human curation.

% \textbf{Level 3: Real-time Interactive Generative Worlds.}
% This level represents a significant advancement where the game engine continuously generates content based on player interactions while adhering to physical rules. In a sandbox game, when players interact with the environment, such as breaking walls or manipulating objects, the engine generates physically accurate responses in real-time. The destruction of a wall, for instance, creates realistic debris and effects based on how the player interacts with it. While many current research work operate at this level~\cite{gamengine,gamefactory,oasis}, there's still room to improve the generalization capabilities of these generative systems.

\textbf{Level 2: Physics-Compliant Interactive World Generation.}
The engine generates physics-compliant content from player interactions in real-time\eg{ (e.g., setting fire to a bridge generates blazing effects and reroutes enemies)}. Many works~\cite{gamengine,gamefactory,oasis} approach this level; improvements are needed in physics understanding and generalization.

% \textbf{Level 4: Persistent Contextual Generative Worlds.}
% At this advanced level, the game engine maintains a sophisticated world model capable of simulating long-term consequences and complex cause-and-effect relationships. In a strategy game, player decisions trigger chains of events that unfold over extended periods. For example, when players deforest an area, the engine simulates the gradual environmental impact over 50 years, generating evolving landscapes that show desertification, changing wildlife patterns, and the emergence of resource conflicts among NPC communities. This deep simulation creates rich, emergent narratives that arise naturally from player actions.

\textbf{Level 3: Causal-Reasoning World Simulation.}
Building on L2, this level adds causal reasoning across time\eg{: assassinating a faction leader in Act 1 triggers political instability and city-wide riots in Act 3}. Players' early choices shape the world's future.

% \textbf{Level 5: Self-Evolving Generative Ecosystem.}
% The highest level represents a fully autonomous game world that can adapt its own rules and continuously generate new content in response to player behavior and system evolution. In a metaverse environment, AI-driven NPCs can develop their own goals and relationships beyond pre-programmed scenarios. For example, NPCs might form groups to trade resources, build new settlements, or change game rules based on their experiences. 
% This level of autonomy creates truly emergent gameplay experiences that transcend traditional game design limitations.

\textbf{Level 4: Self-Evolving World Ecosystem.}
As L2 and L3 advance, complex systems emerge from initial rules\eg{: NPCs autonomously organize governance and trade networks beyond their programming}. Self-evolution remains bounded by designer-defined rules, style, and intervention: distinct games should retain their authored identity, rather than converge to an unconstrained common world.

\section{Alternative Views}
\label{sec:alternative}

\revised{Through discussions with researchers, game developers, and community participants, we have identified a range of alternative views regarding building next-generation game engines on IGV. Below we provide brief responses; full analysis is in Appendix~\ref{appendix:alternative}.}

\revised{\textbf{AV\#1: GGE vs.\ Procedural Content Generation (PCG).}
GGE aims to extend diversity beyond fixed asset combinations by learning content and world priors; generation alone does not guarantee meaningful variety.}

\revised{\textbf{AV\#2: Traditional Rendering vs.\ IGV.}
IGV is not merely a graphics upgrade; it is a complete system (Control, Memory, Dynamics, Intelligence) that enhances both visual quality and gameplay simultaneously.}

\revised{\textbf{AV\#3: Economic Costs.}
Distillation~\cite{causvid} and efficient inference~\cite{cosmos} offer cost-reduction paths, but near-term players may face cloud fees, costly hardware, energy use, and latency. Total costs must be weighed against player value; traditional or hybrid deployment remains appropriate where IGV is unaffordable.}

\revised{\textbf{AV\#4: Video Generation vs.\ 3D Scene Generation.}
Video data is vastly more abundant and expressively unified. We view 3D representations as complementary memory/spatial backbones within IGV, not competing alternatives.}

\revised{\textbf{AV\#5: Traditional Engines + LLM vs.\ IGV.}
LLM-augmented traditional engines face fundamental ceilings in asset combinatorics and rule brittleness. IGV operates at the perception level, generating continuous experiences unconstrained by fixed asset inventories. Our position is that IGV has the higher long-term ceiling.}

\revised{\textbf{AV\#6: Capability $\neq$ Good Engine.}
We distinguish engine capability from game quality: IGV expands the capability boundary, enabling new categories of experiences. Good games remain the product of human creativity using these expanded tools.}

\revised{\textbf{AV\#7: Does IGV Need 100\% Reliability?}
IGV complements rather than replaces traditional physics engines: formula-based physics stays with explicit solvers, while IGV handles formula-hard and semantic physics. Its control value lies in expanding controllable dimensions, not competing on WASD precision.}

\newrevised{\textbf{AV\#8: Hybrid Multi-Technique Composition vs.\ IGV-as-Core.}
Our six-module architecture already accommodates heterogeneous sub-systems (LLM, RL, 3D generation); the position is that IGV serves as the integrating perceptual core, not that it must handle every sub-task alone.}

\textbf{AV\#9: Why Would Players Want an IGV Game?}
Open-ended exploration and personalization may motivate play, but unlimited content is not inherently enjoyable. Designers must preserve goals, pacing, style, and a recognizable game identity. IGV expands their tools; it does not guarantee fun or require players to accept unconstrained generation.

\section{Ethical Issues}
\label{sec:ethical}
Several key ethical issues arise in GGE development: \textbf{copyright} (ownership of AI-generated content), \textbf{security} (preventing harmful content generation), \textbf{creativity} (whether AI enhances or limits human expression), \textbf{democratization} (impact of lowering creation barriers), \textbf{labor} (effects on industry workers), \textbf{environmental impact} (training and runtime energy use), and \textbf{commercial distribution} (platform certification and content rating for generative systems). Candidate mitigations and their limitations are discussed in Appendix~\ref{appendix:ethical}.

\section{Conclusion}
\label{sec:conclusion}
\revised{We have argued that next-generation game engines should be built on Interactive Generative Video. \finalrev{Our contribution is not the observation that video generation can power interactive worlds (which is increasingly recognized), but rather the systematic formalization: a six-module architecture identifying concrete research bottlenecks, a measurable L0--L4 roadmap with observable criteria, and direct engagement with nine competing alternative views.} To support this position, we analyzed why IGV is uniquely suited for this role (Sec.~\ref{sec:why}), identified what is still missing (Sec.~\ref{sec:gap}), surveyed current technical progress across six functional modules (Sec.~\ref{sec:framework}), proposed a staged roadmap with observable criteria (Sec.~\ref{sec:levels}), and engaged with nine alternative views and ethical considerations (Secs.~\ref{sec:alternative}--\ref{sec:ethical}). We believe this comprehensive treatment can stimulate broader community discussion on the future of generative game engines. IGV is not yet mature enough to replace existing engines, and significant technical challenges remain on the path forward, but its potential is substantial. Whether or not IGV ultimately becomes the core of next-generation engines, the open problems identified in this paper require joint effort from both the research community and the game industry to address.}

% \section*{Impact Statement}
% This paper proposes Interactive Generative Videos (IGV) as a core technology for the next-generation game engine, Generative Game Engine (GGE). IGV will transform the way generative AI is utilized, further advancing the development of interactive applications. GGE will revolutionize the gaming industry by enabling rapid production of high-quality, unlimited, and adaptive game content.

\begin{ack}
This work is supported by the Research Grant Council of Hong Kong through Early Career Scheme under grant 27214925.
\end{ack}

\bibliographystyle{plain}
\bibliography{example}

@article{ddpm,
  title={Denoising diffusion probabilistic models},
  author={Ho, Jonathan and Jain, Ajay and Abbeel, Pieter},
  journal={Advances in neural information processing systems},
  year={2020}
}

@article{sde,
  title={Score-based generative modeling through stochastic differential equations},
  author={Song, Yang and Sohl-Dickstein, Jascha and Kingma, Diederik P and Kumar, Abhishek and Ermon, Stefano and Poole, Ben},
  journal={International Conference on Learning Representations},
  year={2021}
}

@article{score,
  title={Generative modeling by estimating gradients of the data distribution},
  author={Song, Yang and Ermon, Stefano},
  journal={Advances in neural information processing systems},
  year={2019}
}

@inproceedings{sd,
  title={High-resolution image synthesis with latent diffusion models},
  author={Rombach, Robin and Blattmann, Andreas and Lorenz, Dominik and Esser, Patrick and Ommer, Bj{\"o}rn},
  booktitle={Proceedings of the IEEE/CVF conference on computer vision and pattern recognition},
  year={2022}
}

@misc{sora,
  author = {OpenAI},
  title = {Creating video from text},
  howpublished = {\url{https://openai.com/index/sora/}},
  year = {2024}, 
}

@article{cogvideox,
  title={CogVideoX: Text-to-Video Diffusion Models with An Expert Transformer},
  author={Yang, Zhuoyi and Teng, Jiayan and Zheng, Wendi and Ding, Ming and Huang, Shiyu and Xu, Jiazheng and Yang, Yuanming and Hong, Wenyi and Zhang, Xiaohan and Feng, Guanyu and others},
  journal={arXiv preprint arXiv:2408.06072},
  year={2024}
}

@inproceedings{genie,
  title={Genie: Generative interactive environments},
  author={Bruce, Jake and Dennis, Michael D and Edwards, Ashley and Parker-Holder, Jack and Shi, Yuge and Hughes, Edward and Lai, Matthew and Mavalankar, Aditi and Steigerwald, Richie and Apps, Chris and others},
  booktitle={Forty-first International Conference on Machine Learning},
  year={2024}
}

@article{gamengine,
  title={Diffusion models are real-time game engines},
  author={Valevski, Dani and Leviathan, Yaniv and Arar, Moab and Fruchter, Shlomi},
  journal={arXiv preprint arXiv:2408.14837},
  year={2024}
}

@inproceedings{diamond,
      title={Diffusion for World Modeling: Visual Details Matter in Atari},
      author={Eloi Alonso and Adam Jelley and Vincent Micheli and Anssi Kanervisto and Amos Storkey and Tim Pearce and François Fleuret},
      booktitle={Thirty-eighth Conference on Neural Information Processing Systems},
      year={2024}
}

@misc{oasis,
  author = {Decart, Etched},
  title = {Oasis: A Universe in a Transformer},
  howpublished = {\url{https://oasis-model.github.io/}},
  year = {2024}, 
}

@article{gamegenx,
  title={Gamegen-x: Interactive open-world game video generation},
  author={Che, Haoxuan and He, Xuanhua and Liu, Quande and Jin, Cheng and Chen, Hao},
  journal={arXiv preprint arXiv:2411.00769},
  year={2024}
}

@article{unisim,
title={Learning Interactive Real-World Simulators},
author={Yang, Mengjiao and Du, Yilun and Ghasemipour, Kamyar and Tompson, Jonathan and Schuurmans, Dale and Abbeel, Pieter},
journal={arXiv preprint arXiv:2310.06114},
year={2023}
}

@InProceedings{video-new-language,
  title = 	 {Position: Video as the New Language for Real-World Decision Making},
  author =       {Yang, Sherry and Walker, Jacob C and Parker-Holder, Jack and Du, Yilun and Bruce, Jake and Barreto, Andre and Abbeel, Pieter and Schuurmans, Dale},
  booktitle = 	 {Proceedings of the 41st International Conference on Machine Learning},
  year = 	 {2024}
}

@inproceedings{motionctrl,
  title={Motionctrl: A unified and flexible motion controller for video generation},
  author={Wang, Zhouxia and Yuan, Ziyang and Wang, Xintao and Li, Yaowei and Chen, Tianshui and Xia, Menghan and Luo, Ping and Shan, Ying},
  booktitle={ACM SIGGRAPH 2024 Conference Papers},
  year={2024}
}

@misc{dynamicrafter,
      title={DynamiCrafter: Animating Open-domain Images with Video Diffusion Priors}, 
      author={Jinbo Xing and Menghan Xia and Yong Zhang and Haoxin Chen and Xintao Wang and Tien-Tsin Wong and Ying Shan},
      year={2023},
      eprint={2310.12190},
      archivePrefix={arXiv}
}

@inproceedings{minedojo,
  title     = {MineDojo: Building Open-Ended Embodied Agents with Internet-Scale Knowledge},
  author    = {Linxi Fan and Guanzhi Wang and Yunfan Jiang and Ajay Mandlekar and Yuncong Yang and Haoyi Zhu and Andrew Tang and De-An Huang and Yuke Zhu and Anima Anandkumar},
  booktitle = {Thirty-sixth Conference on Neural Information Processing Systems Datasets and Benchmarks Track},
  year      = {2022},
}

@article{cameractrl,
  title={Cameractrl: Enabling camera control for text-to-video generation},
  author={He, Hao and Xu, Yinghao and Guo, Yuwei and Wetzstein, Gordon and Dai, Bo and Li, Hongsheng and Yang, Ceyuan},
  journal={arXiv preprint arXiv:2404.02101},
  year={2024}
}

@inproceedings{guo2025sparsectrl,
  title={Sparsectrl: Adding sparse controls to text-to-video diffusion models},
  author={Guo, Yuwei and Yang, Ceyuan and Rao, Anyi and Agrawala, Maneesh and Lin, Dahua and Dai, Bo},
  booktitle={European Conference on Computer Vision},
  pages={330--348},
  year={2025},
  organization={Springer}
}

@inproceedings{ni2023conditional,
  title={Conditional image-to-video generation with latent flow diffusion models},
  author={Ni, Haomiao and Shi, Changhao and Li, Kai and Huang, Sharon X and Min, Martin Renqiang},
  booktitle={Proceedings of the IEEE/CVF conference on computer vision and pattern recognition},
  pages={18444--18455},
  year={2023}
}

@article{worldsimbench,
  title={Worldsimbench: Towards video generation models as world simulators},
  author={Qin, Yiran and Shi, Zhelun and Yu, Jiwen and Wang, Xijun and Zhou, Enshen and Li, Lijun and Yin, Zhenfei and Liu, Xihui and Sheng, Lu and Shao, Jing and others},
  journal={arXiv preprint arXiv:2410.18072},
  year={2024}
}

@inproceedings{mp5,
  title={MP5: A Multi-modal Open-ended Embodied System in Minecraft via Active Perception},
  author={Qin, Yiran and Zhou, Enshen and Liu, Qichang and Yin, Zhenfei and Sheng, Lu and Zhang, Ruimao and Qiao, Yu and Shao, Jing},
  booktitle={Proceedings of the IEEE/CVF Conference on Computer Vision and Pattern Recognition},
  pages={16307--16316},
  year={2024}
}

@article{minedreamer,
  title={Minedreamer: Learning to follow instructions via chain-of-imagination for simulated-world control},
  author={Zhou, Enshen and Qin, Yiran and Yin, Zhenfei and Huang, Yuzhou and Zhang, Ruimao and Sheng, Lu and Qiao, Yu and Shao, Jing},
  journal={arXiv preprint arXiv:2403.12037},
  year={2024}
}

@article{matrix,
  title={The Matrix: Infinite-Horizon World Generation with Real-Time Moving Control},
  author={Feng, Ruili and Zhang, Han and Yang, Zhantao and Xiao, Jie and Shu, Zhilei and Liu, Zhiheng and Zheng, Andy and Huang, Yukun and Liu, Yu and Zhang, Hongyang},
  journal={arXiv preprint arXiv:2412.03568},
  year={2024}
}

@article{playgen,
  title={Playable Game Generation},
  author={Yang, Mingyu and Li, Junyou and Fang, Zhongbin and Chen, Sheng and Yu, Yangbin and Fu, Qiang and Yang, Wei and Ye, Deheng},
  journal={arXiv preprint arXiv:2412.00887},
  year={2024}
}

@misc{genie2,
  author = {Google DeepMind},
  title = {Genie 2: A large-scale foundation world model},
  howpublished = {https://deepmind.google/discover/blog/genie-2-a-large-scale-foundation-world-model/},
  year = {2024}, 
}

@article{videopoet,
  title={Videopoet: A large language model for zero-shot video generation},
  author={Kondratyuk, Dan and Yu, Lijun and Gu, Xiuye and Lezama, Jos{\'e} and Huang, Jonathan and Schindler, Grant and Hornung, Rachel and Birodkar, Vighnesh and Yan, Jimmy and Chiu, Ming-Chang and others},
  journal={arXiv preprint arXiv:2312.14125},
  year={2023}
}

@article{videogpt,
  title={Videogpt: Video generation using vq-vae and transformers},
  author={Yan, Wilson and Zhang, Yunzhi and Abbeel, Pieter and Srinivas, Aravind},
  journal={arXiv preprint arXiv:2104.10157},
  year={2021}
}

@article{cosmos,
  title={Cosmos World Foundation Model Platform for Physical AI},
  author={NVIDIA},
  journal={arXiv preprint arXiv:2501.03575},
  year={2025}
}

@misc{gamefactory,
      title={GameFactory: Creating New Games with Generative Interactive Videos}, 
      author={Yu, Jiwen and Qin, Yiran and Wang, Xintao and Wan, Pengfei and Zhang, Di and Liu, Xihui},
      year={2025},
      eprint={2501.08325},
      archivePrefix={arXiv},
}

@article{nova,
  title={Autoregressive Video Generation without Vector Quantization},
  author={Deng, Haoge and Pan, Ting and Diao, Haiwen and Luo, Zhengxiong and Cui, Yufeng and Lu, Huchuan and Shan, Shiguang and Qi, Yonggang and Wang, Xinlong},
  journal={arXiv preprint arXiv:2412.14169},
  year={2024}
}

@article{cdf,
  title={Diffusion forcing: Next-token prediction meets full-sequence diffusion},
  author={Chen, Boyuan and Monso, Diego Marti and Du, Yilun and Simchowitz, Max and Tedrake, Russ and Sitzmann, Vincent},
  journal={arXiv preprint arXiv:2407.01392},
  year={2024}
}

@article{emu3,
  title={Emu3: Next-token prediction is all you need},
  author={Wang, Xinlong and Zhang, Xiaosong and Luo, Zhengxiong and Sun, Quan and Cui, Yufeng and Wang, Jinsheng and Zhang, Fan and Wang, Yueze and Li, Zhen and Yu, Qiying and others},
  journal={arXiv preprint arXiv:2409.18869},
  year={2024}
}

@article{zhou2024transfusion,
  title={Transfusion: Predict the next token and diffuse images with one multi-modal model},
  author={Zhou, Chunting and Yu, Lili and Babu, Arun and Tirumala, Kushal and Yasunaga, Michihiro and Shamis, Leonid and Kahn, Jacob and Ma, Xuezhe and Zettlemoyer, Luke and Levy, Omer},
  journal={arXiv preprint arXiv:2408.11039},
  year={2024}
}

@article{xie2024show,
  title={Show-o: One single transformer to unify multimodal understanding and generation},
  author={Xie, Jinheng and Mao, Weijia and Bai, Zechen and Zhang, David Junhao and Wang, Weihao and Lin, Kevin Qinghong and Gu, Yuchao and Chen, Zhijie and Yang, Zhenheng and Shou, Mike Zheng},
  journal={arXiv preprint arXiv:2408.12528},
  year={2024}
}

@inproceedings{3dtrajmaster,
  title={3DTrajMaster: Mastering 3D Trajectory for Multi-Entity Motion in Video Generation},
  author={Fu, Xiao and Liu, Xian and Wang, Xintao and Peng, Sida and Xia, Menghan and Shi, Xiaoyu and Yuan, Ziyang and Wan, Pengfei and Zhang, Di and Lin, Dahua},
  booktitle={ICLR},
  year={2025}
}

@inproceedings{direct-a-video,
  title={Direct-a-video: Customized video generation with user-directed camera movement and object motion},
  author={Yang, Shiyuan and Hou, Liang and Huang, Haibin and Ma, Chongyang and Wan, Pengfei and Zhang, Di and Chen, Xiaodong and Liao, Jing},
  booktitle={ACM SIGGRAPH 2024 Conference Papers},
  pages={1--12},
  year={2024}
}

@article{razavi2019generating,
  title={Generating diverse high-fidelity images with vq-vae-2},
  author={Razavi, Ali and Van den Oord, Aaron and Vinyals, Oriol},
  journal={Advances in neural information processing systems},
  volume={32},
  year={2019}
}

@article{imagen,
  title={Imagen video: High definition video generation with diffusion models},
  author={Ho, Jonathan and Chan, William and Saharia, Chitwan and Whang, Jay and Gao, Ruiqi and Gritsenko, Alexey and Kingma, Diederik P and Poole, Ben and Norouzi, Mohammad and Fleet, David J and others},
  journal={arXiv preprint arXiv:2210.02303},
  year={2022}
}

@inproceedings{maskgit,
  title={Maskgit: Masked generative image transformer},
  author={Chang, Huiwen and Zhang, Han and Jiang, Lu and Liu, Ce and Freeman, William T},
  booktitle={Proceedings of the IEEE/CVF Conference on Computer Vision and Pattern Recognition},
  pages={11315--11325},
  year={2022}
}

@inproceedings{gamegan,
  title={Learning to simulate dynamic environments with gamegan},
  author={Kim, Seung Wook and Zhou, Yuhao and Philion, Jonah and Torralba, Antonio and Fidler, Sanja},
  booktitle={Proceedings of the IEEE/CVF Conference on Computer Vision and Pattern Recognition},
  pages={1231--1240},
  year={2020}
}

@inproceedings{caddy,
  title={Playable video generation},
  author={Menapace, Willi and Lathuiliere, Stephane and Tulyakov, Sergey and Siarohin, Aliaksandr and Ricci, Elisa},
  booktitle={Proceedings of the IEEE/CVF Conference on Computer Vision and Pattern Recognition},
  pages={10061--10070},
  year={2021}
}

@inproceedings{drivegan,
  title={Drivegan: Towards a controllable high-quality neural simulation},
  author={Kim, Seung Wook and Philion, Jonah and Torralba, Antonio and Fidler, Sanja},
  booktitle={Proceedings of the IEEE/CVF Conference on Computer Vision and Pattern Recognition},
  pages={5820--5829},
  year={2021}
}

@inproceedings{pe,
  title={Playable environments: Video manipulation in space and time},
  author={Menapace, Willi and Lathuili{\`e}re, St{\'e}phane and Siarohin, Aliaksandr and Theobalt, Christian and Tulyakov, Sergey and Golyanik, Vladislav and Ricci, Elisa},
  booktitle={Proceedings of the IEEE/CVF Conference on Computer Vision and Pattern Recognition},
  pages={3584--3593},
  year={2022}
}

@inproceedings{gsb,
  title={Open-world multi-task control through goal-aware representation learning and adaptive horizon prediction},
  author={Cai, Shaofei and Wang, Zihao and Ma, Xiaojian and Liu, Anji and Liang, Yitao},
  booktitle={Proceedings of the IEEE/CVF Conference on Computer Vision and Pattern Recognition},
  pages={13734--13744},
  year={2023}
}

@inproceedings{clip4mc,
  title={Reinforcement learning friendly vision-language model for minecraft},
  author={Jiang, Haobin and Yue, Junpeng and Luo, Hao and Ding, Ziluo and Lu, Zongqing},
  booktitle={European Conference on Computer Vision},
  pages={1--17},
  year={2025},
  organization={Springer}
}

@article{minerl,
  title={Minerl: A large-scale dataset of minecraft demonstrations},
  author={Guss, William H and Houghton, Brandon and Topin, Nicholay and Wang, Phillip and Codel, Cayden and Veloso, Manuela and Salakhutdinov, Ruslan},
  journal={arXiv preprint arXiv:1907.13440},
  year={2019}
}

@article{juewu,
  title={Juewu-mc: Playing minecraft with sample-efficient hierarchical reinforcement learning},
  author={Lin, Zichuan and Li, Junyou and Shi, Jianing and Ye, Deheng and Fu, Qiang and Yang, Wei},
  journal={arXiv preprint arXiv:2112.04907},
  year={2021}
}

@inproceedings{zerorl,
  title={Zero-shot task generalization with multi-task deep reinforcement learning},
  author={Oh, Junhyuk and Singh, Satinder and Lee, Honglak and Kohli, Pushmeet},
  booktitle={International Conference on Machine Learning},
  pages={2661--2670},
  year={2017},
  organization={PMLR}
}

@article{plan4mc,
  title={Plan4mc: Skill reinforcement learning and planning for open-world minecraft tasks},
  author={BAAI, PKU},
  journal={arXiv preprint arXiv:2303.16563},
  year={2023}
}

@article{voyager,
  title={Voyager: An Open-Ended Embodied Agent with Large Language Models. 2023},
  author={Wang, Guanzhi and Xie, Yuqi and Jiang, Yunfan and Mandlekar, Ajay and Xiao, Chaowei and Zhu, Yuke and Fan, Linxi and Anandkumar, Anima},
  journal={Comment: Project website and open-source codebase: https://voyager. minedojo. org/Cited on},
  pages={33},
  year={2023}
}

@article{deps,
  title={Describe, explain, plan and select: Interactive planning with large language models enables open-world multi-task agents},
  author={Wang, Zihao and Cai, Shaofei and Chen, Guanzhou and Liu, Anji and Ma, Xiaojian and Liang, Yitao},
  journal={arXiv preprint arXiv:2302.01560},
  year={2023}
}

@article{gitm,
  title={Ghost in the minecraft: Generally capable agents for open-world environments via large language models with text-based knowledge and memory},
  author={Zhu, Xizhou and Chen, Yuntao and Tian, Hao and Tao, Chenxin and Su, Weijie and Yang, Chenyu and Huang, Gao and Li, Bin and Lu, Lewei and Wang, Xiaogang and others},
  journal={arXiv preprint arXiv:2305.17144},
  year={2023}
}

@inproceedings{llmplanner,
  title={Language models as zero-shot planners: Extracting actionable knowledge for embodied agents},
  author={Huang, Wenlong and Abbeel, Pieter and Pathak, Deepak and Mordatch, Igor},
  booktitle={International conference on machine learning},
  pages={9118--9147},
  year={2022},
  organization={PMLR}
}

@article{llama,
  title={Llama: Open and efficient foundation language models},
  author={Touvron, Hugo and Lavril, Thibaut and Izacard, Gautier and Martinet, Xavier and Lachaux, Marie-Anne and Lacroix, Timoth{\'e}e and Rozi{\`e}re, Baptiste and Goyal, Naman and Hambro, Eric and Azhar, Faisal and others},
  journal={arXiv preprint arXiv:2302.13971},
  year={2023}
}

@article{gpt4,
  title={Gpt-4 technical report},
  author={Achiam, Josh and Adler, Steven and Agarwal, Sandhini and Ahmad, Lama and Akkaya, Ilge and Aleman, Florencia Leoni and Almeida, Diogo and Altenschmidt, Janko and Altman, Sam and Anadkat, Shyamal and others},
  journal={arXiv preprint arXiv:2303.08774},
  year={2023}
}

@inproceedings{octopus,
  title={Octopus: Embodied vision-language programmer from environmental feedback},
  author={Yang, Jingkang and Dong, Yuhao and Liu, Shuai and Li, Bo and Wang, Ziyue and Tan, Haoran and Jiang, Chencheng and Kang, Jiamu and Zhang, Yuanhan and Zhou, Kaiyang and others},
  booktitle={European Conference on Computer Vision},
  pages={20--38},
  year={2025},
  organization={Springer}
}

@article{causvid,
  title={From slow bidirectional to fast causal video generators},
  author={Yin, Tianwei and Zhang, Qiang and Zhang, Richard and Freeman, William T and Durand, Fredo and Shechtman, Eli and Huang, Xun},
  journal={arXiv preprint arXiv:2412.07772},
  year={2024}
}

@article{dcae,
  title={Deep compression autoencoder for efficient high-resolution diffusion models},
  author={Chen, Junyu and Cai, Han and Chen, Junsong and Xie, Enze and Yang, Shang and Tang, Haotian and Li, Muyang and Lu, Yao and Han, Song},
  journal={arXiv preprint arXiv:2410.10733},
  year={2024}
}

@article{videolcm,
  title={Videolcm: Video latent consistency model},
  author={Wang, Xiang and Zhang, Shiwei and Zhang, Han and Liu, Yu and Zhang, Yingya and Gao, Changxin and Sang, Nong},
  journal={arXiv preprint arXiv:2312.09109},
  year={2023}
}

@inproceedings{bk-sdm,
  title={Bk-sdm: A lightweight, fast, and cheap version of stable diffusion},
  author={Kim, Bo-Kyeong and Song, Hyoung-Kyu and Castells, Thibault and Choi, Shinkook},
  booktitle={European Conference on Computer Vision},
  pages={381--399},
  year={2025},
  organization={Springer}
}

@inproceedings{pengi,
 author = {Deshmukh, Soham and Elizalde, Benjamin and Singh, Rita and Wang, Huaming},
 booktitle = {Advances in Neural Information Processing Systems},
 title = {Pengi: An Audio Language Model for Audio Tasks},
 year = {2023}
}

@inproceedings{depthanything,
  title={Depth anything: Unleashing the power of large-scale unlabeled data},
  author={Yang, Lihe and Kang, Bingyi and Huang, Zilong and Xu, Xiaogang and Feng, Jiashi and Zhao, Hengshuang},
  booktitle={Proceedings of the IEEE/CVF Conference on Computer Vision and Pattern Recognition},
  pages={10371--10381},
  year={2024}
}

@article{motiongpt,
  title={Motiongpt: Human motion as a foreign language},
  author={Jiang, Biao and Chen, Xin and Liu, Wen and Yu, Jingyi and Yu, Gang and Chen, Tao},
  journal={Advances in Neural Information Processing Systems},
  volume={36},
  pages={20067--20079},
  year={2023}
}

@inproceedings{lvm,
  title={Sequential modeling enables scalable learning for large vision models},
  author={Bai, Yutong and Geng, Xinyang and Mangalam, Karttikeya and Bar, Amir and Yuille, Alan L and Darrell, Trevor and Malik, Jitendra and Efros, Alexei A},
  booktitle={Proceedings of the IEEE/CVF Conference on Computer Vision and Pattern Recognition},
  pages={22861--22872},
  year={2024}
}

@article{pgm,
  title={Promptable game models: Text-guided game simulation via masked diffusion models},
  author={Menapace, Willi and Siarohin, Aliaksandr and Lathuili{\`e}re, St{\'e}phane and Achlioptas, Panos and Golyanik, Vladislav and Tulyakov, Sergey and Ricci, Elisa},
  journal={ACM Transactions on Graphics},
  volume={43},
  number={2},
  pages={1--16},
  year={2024},
  publisher={ACM New York, NY, USA}
}

@inproceedings{wonderjourney,
  title={Wonderjourney: Going from anywhere to everywhere},
  author={Yu, Hong-Xing and Duan, Haoyi and Hur, Junhwa and Sargent, Kyle and Rubinstein, Michael and Freeman, William T and Cole, Forrester and Sun, Deqing and Snavely, Noah and Wu, Jiajun and others},
  booktitle={Proceedings of the IEEE/CVF Conference on Computer Vision and Pattern Recognition},
  pages={6658--6667},
  year={2024}
}

@article{viewcrafter,
  title={Viewcrafter: Taming video diffusion models for high-fidelity novel view synthesis},
  author={Yu, Wangbo and Xing, Jinbo and Yuan, Li and Hu, Wenbo and Li, Xiaoyu and Huang, Zhipeng and Gao, Xiangjun and Wong, Tien-Tsin and Shan, Ying and Tian, Yonghong},
  journal={arXiv preprint arXiv:2409.02048},
  year={2024}
}

@phdthesis{liapis2015searching,
    title={Searching for sentient design tools for game development},
    author={Liapis, Antonios},
    year={2015},
    school={IT University of Copenhagen},
    type={PhD Thesis}
}

@article{tanagra,
  title={Tanagra: Reactive planning and constraint solving for mixed-initiative level design},
  author={Smith, Gillian and Whitehead, Jim and Mateas, Michael},
  journal={IEEE Transactions on computational intelligence and AI in games},
  volume={3},
  number={3},
  pages={201--215},
  year={2011},
  publisher={IEEE}
}

@inproceedings{sketchbook,
  title={Designer modeling for sentient sketchbook},
  author={Liapis, Antonios and Yannakakis, Georgios N and Togelius, Julian},
  booktitle={2014 IEEE Conference on Computational Intelligence and Games},
  pages={1--8},
  year={2014},
  organization={IEEE}
}

@article{susketch,
  title={SuSketch: Surrogate models of gameplay as a design assistant},
  author={Migkotzidis, Panagiotis and Liapis, Antonios},
  journal={IEEE Transactions on Games},
  volume={14},
  number={2},
  pages={273--283},
  year={2021},
  publisher={IEEE}
}

@article{baba,
  title={Baba is y'all 2.0: Design and investigation of a collaborative mixed-initiative system},
  author={Charity, Megan and Dave, Isha and Khalifa, Ahmed and Togelius, Julian},
  journal={IEEE Transactions on Games},
  volume={16},
  number={1},
  pages={75--89},
  year={2022},
  publisher={IEEE}
}

@inproceedings{lottery,
  title={Lottery and sprint: Generate a board game with design sprint method on AutoGPT},
  author={Torii, Maya Grace and Murakami, Takahito and Ochiai, Yoichi},
  booktitle={Companion Proceedings of the Annual Symposium on Computer-Human Interaction in Play},
  pages={259--265},
  year={2023}
}

@article{see3d,
  title={You See it, You Got it: Learning 3D Creation on Pose-Free Videos at Scale},
  author={Ma, Baorui and Gao, Huachen and Deng, Haoge and Luo, Zhengxiong and Huang, Tiejun and Tang, Lulu and Wang, Xinlong},
  journal={arXiv preprint arXiv:2412.06699},
  year={2024}
}

@article{vid2sim,
  title={Vid2Sim: Realistic and Interactive Simulation from Video for Urban Navigation},
  author={Xie, Ziyang and Liu, Zhizheng and Peng, Zhenghao and Wu, Wayne and Zhou, Bolei},
  journal={arXiv preprint arXiv:2501.06693},
  year={2025}
}

@article{how-far-world-model,
  title={How far is video generation from world model: A physical law perspective},
  author={Kang, Bingyi and Yue, Yang and Lu, Rui and Lin, Zhijie and Zhao, Yang and Wang, Kaixin and Huang, Gao and Feng, Jiashi},
  journal={arXiv preprint arXiv:2411.02385},
  year={2024}
}

@article{Physics-IQ,
  title={Do generative video models learn physical principles from watching videos?},
  author={Motamed, Saman and Culp, Laura and Swersky, Kevin and Jaini, Priyank and Geirhos, Robert},
  journal={arXiv preprint arXiv:2501.09038},
  year={2025}
}

@article{llava,
  title={Visual instruction tuning},
  author={Liu, Haotian and Li, Chunyuan and Wu, Qingyang and Lee, Yong Jae},
  journal={Advances in neural information processing systems},
  volume={36},
  year={2024}
}

@article{camel,
  title={Camel: Communicative agents for" mind" exploration of large language model society},
  author={Li, Guohao and Hammoud, Hasan and Itani, Hani and Khizbullin, Dmitrii and Ghanem, Bernard},
  journal={Advances in Neural Information Processing Systems},
  volume={36},
  pages={51991--52008},
  year={2023}
}

@article{autogen,
  title={Autogen: Enabling next-gen llm applications via multi-agent conversation framework},
  author={Wu, Qingyun and Bansal, Gagan and Zhang, Jieyu and Wu, Yiran and Zhang, Shaokun and Zhu, Erkang and Li, Beibin and Jiang, Li and Zhang, Xiaoyun and Wang, Chi},
  journal={arXiv preprint arXiv:2308.08155},
  year={2023}
}

@article{mar,
  title={Autoregressive Image Generation without Vector Quantization},
  author={Li, Tianhong and Tian, Yonglong and Li, He and Deng, Mingyang and He, Kaiming},
  journal={arXiv preprint arXiv:2406.11838},
  year={2024}
}

@article{gato,
  title={A generalist agent},
  author={Reed, Scott and Zolna, Konrad and Parisotto, Emilio and Colmenarejo, Sergio Gomez and Novikov, Alexander and Barth-Maron, Gabriel and Gimenez, Mai and Sulsky, Yury and Kay, Jackie and Springenberg, Jost Tobias and others},
  journal={arXiv preprint arXiv:2205.06175},
  year={2022}
}

@article{domainwatermark,
  title={Domain watermark: Effective and harmless dataset copyright protection is closed at hand},
  author={Guo, Junfeng and Li, Yiming and Wang, Lixu and Xia, Shu-Tao and Huang, Heng and Liu, Cong and Li, Bo},
  journal={Advances in Neural Information Processing Systems},
  volume={36},
  pages={54421--54450},
  year={2023}
}

@article{zeromark,
  title={Zeromark: Towards dataset ownership verification without disclosing watermark},
  author={Guo, Junfeng and Li, Yiming and Chen, Ruibo and Wu, Yihan and Huang, Heng and others},
  journal={Advances in Neural Information Processing Systems},
  volume={37},
  pages={120468--120500},
  year={2024}
}

@article{safesora,
  title={Safesora: Towards safety alignment of text2video generation via a human preference dataset},
  author={Dai, Juntao and Chen, Tianle and Wang, Xuyao and Yang, Ziran and Chen, Taiye and Ji, Jiaming and Yang, Yaodong},
  journal={Advances in Neural Information Processing Systems},
  volume={37},
  pages={17161--17214},
  year={2024}
}

@inproceedings{safeld,
  title={Safe latent diffusion: Mitigating inappropriate degeneration in diffusion models},
  author={Schramowski, Patrick and Brack, Manuel and Deiseroth, Bj{\"o}rn and Kersting, Kristian},
  booktitle={Proceedings of the IEEE/CVF Conference on Computer Vision and Pattern Recognition},
  pages={22522--22531},
  year={2023}
}

@article{mllmasjudge,
  title={MLLM-as-a-Judge for Image Safety without Human Labeling},
  author={Wang, Zhenting and Hu, Shuming and Zhao, Shiyu and Lin, Xiaowen and Juefei-Xu, Felix and Li, Zhuowei and Han, Ligong and Subramanyam, Harihar and Chen, Li and Chen, Jianfa and others},
  journal={arXiv preprint arXiv:2501.00192},
  year={2024}
}

@inproceedings{llavaguard,
  title={LLAVAGUARD: VLM-based Safeguard for Vision Dataset Curation and Safety Assessment},
  author={Helff, Lukas and Friedrich, Felix and Brack, Manuel and Schramowski, Patrick and Kersting, Kristian},
  booktitle={Proceedings of the IEEE/CVF Conference on Computer Vision and Pattern Recognition},
  pages={8322--8326},
  year={2024}
}

@article{zhao2025synthetic,
  title={Synthetic video enhances physical fidelity in video synthesis},
  author={Zhao, Qi and Ni, Xingyu and Wang, Ziyu and Cheng, Feng and Yang, Ziyan and Jiang, Lu and Wang, Bohan},
  journal={arXiv preprint arXiv:2503.20822},
  year={2025}
}

@misc{kling,
  author = {Kling},
  title = {Kling AI: Next-Generation AI Creative Studio},
  howpublished = {\url{https://app.klingai.com/}},
  year = {2024}, 
}

@misc{veo2,
  author = {Google DeepMind},
  title = {Veo 2: Our state-of-the-art video generation model},
  howpublished = {\url{https://deepmind.google/technologies/veo/veo-2/}},
  year = {2024}, 
}

@article{seaweed,
  title={Seaweed-7B: Cost-Effective Training of Video Generation Foundation Model},
  author={Seawead, Team and Yang, Ceyuan and Lin, Zhijie and Zhao, Yang and Lin, Shanchuan and Ma, Zhibei and Guo, Haoyuan and Chen, Hao and Qi, Lu and Wang, Sen and others},
  journal={arXiv preprint arXiv:2504.08685},
  year={2025}
}

@article{moviegen,
  title={Movie Gen: A Cast of Media Foundation Models},
  author={The Movie Gen team},
  journal={arXiv preprint arXiv:2410.13720},
  year={2024}
}

@misc{runway,
  author = {Runway},
  title = {Runway : Tools for human imagination.},
  howpublished = {\url{https://runwayml.com/}},
  year = {2024}, 
}

@misc{chatgpt,
  author = {OpenAI},
  title = {Introducing ChatGPT},
  howpublished = {\url{https://openai.com/index/chatgpt/}},
  year = {2022}, 
}

@misc{gpt-4o-image-generation,
  author = {OpenAI},
  title = {Introducing 4o Image Generation},
  howpublished = {\url{https://openai.com/index/introducing-4o-image-generation/}},
  year = {2022}, 
}

@misc{midjourney,
  author = {Midjourney},
  title = {Midjourney},
  howpublished = {\url{https://www.midjourney.com/home}},
  year = {2022}, 
}

@misc{luma,
  author = {Luma AI},
  title = {Luma AI},
  howpublished = {\url{https://lumalabs.ai/}},
  year = {2024}, 
}

@misc{civitai,
  author = {Civitai},
  title = {Civitai},
  howpublished = {\url{https://civitai.com/}},
  year = {2022}, 
}

@article{vidu,
  title={Vidu: a highly consistent, dynamic and skilled text-to-video generator with diffusion models},
  author={Bao, Fan and Xiang, Chendong and Yue, Gang and He, Guande and Zhu, Hongzhou and Zheng, Kaiwen and Zhao, Min and Liu, Shilong and Wang, Yaole and Zhu, Jun},
  journal={arXiv preprint arXiv:2405.04233},
  year={2024}
}

@article{wan,
  title={Wan: Open and Advanced Large-Scale Video Generative Models},
  author={Wang, Ang and Ai, Baole and Wen, Bin and Mao, Chaojie and Xie, Chen-Wei and Chen, Di and Yu, Feiwu and Zhao, Haiming and Yang, Jianxiao and Zeng, Jianyuan and others},
  journal={arXiv preprint arXiv:2503.20314},
  year={2025}
}

@article{hunyuanvideo,
  title={Hunyuanvideo: A systematic framework for large video generative models},
  author={Kong, Weijie and Tian, Qi and Zhang, Zijian and Min, Rox and Dai, Zuozhuo and Zhou, Jin and Xiong, Jiangfeng and Li, Xin and Wu, Bo and Zhang, Jianwei and others},
  journal={arXiv preprint arXiv:2412.03603},
  year={2024}
}

@article{wham,
  title={World and human action models towards gameplay ideation},
  author={Kanervisto, Anssi and Bignell, Dave and Wen, Linda Yilin and Grayson, Martin and Georgescu, Raluca and Valcarcel Macua, Sergio and Tan, Shan Zheng and Rashid, Tabish and Pearce, Tim and Cao, Yuhan and others},
  journal={Nature},
  volume={638},
  number={8051},
  pages={656--663},
  year={2025},
  publisher={Nature Publishing Group UK London}
}

@article{mineworld,
  title={MineWorld: a Real-Time and Open-Source Interactive World Model on Minecraft},
  author={Guo, Junliang and Ye, Yang and He, Tianyu and Wu, Haoyu and Jiang, Yushu and Pearce, Tim and Bian, Jiang},
  journal={arXiv preprint arXiv:2504.08388},
  year={2025}
}

@article{adaworld,
  title={AdaWorld: Learning Adaptable World Models with Latent Actions},
  author={Gao, Shenyuan and Zhou, Siyuan and Du, Yilun and Zhang, Jun and Gan, Chuang},
  journal={arXiv preprint arXiv:2503.18938},
  year={2025}
}

@article{maag,
  title={Model as a Game: On Numerical and Spatial Consistency for Generative Games},
  author={Chen, Jingye and Zhao, Yuzhong and Huang, Yupan and Cui, Lei and Dong, Li and Lv, Tengchao and Chen, Qifeng and Wei, Furu},
  journal={arXiv preprint arXiv:2503.21172},
  year={2025}
}

@article{worldmem,
  title={WORLDMEM: Long-term Consistent World Simulation with Memory},
  author={Xiao, Zeqi and Lan, Yushi and Zhou, Yifan and Ouyang, Wenqi and Yang, Shuai and Zeng, Yanhong and Pan, Xingang},
  journal={arXiv preprint arXiv:2504.12369},
  year={2025}
}

@article{flow,
  title={Flow matching for generative modeling},
  author={Lipman, Yaron and Chen, Ricky TQ and Ben-Hamu, Heli and Nickel, Maximilian and Le, Matt},
  journal={arXiv preprint arXiv:2210.02747},
  year={2022}
}

@article{rectified,
  title={Flow straight and fast: Learning to generate and transfer data with rectified flow},
  author={Liu, Xingchao and Gong, Chengyue and Liu, Qiang},
  journal={arXiv preprint arXiv:2209.03003},
  year={2022}
}

@inproceedings{magvit,
  title={Magvit: Masked generative video transformer},
  author={Yu, Lijun and Cheng, Yong and Sohn, Kihyuk and Lezama, Jos{\'e} and Zhang, Han and Chang, Huiwen and Hauptmann, Alexander G and Yang, Ming-Hsuan and Hao, Yuan and Essa, Irfan and others},
  booktitle={Proceedings of the IEEE/CVF Conference on Computer Vision and Pattern Recognition},
  pages={10459--10469},
  year={2023}
}

@article{recammaster,
  title={ReCamMaster: Camera-Controlled Generative Rendering from A Single Video},
  author={Bai, Jianhong and Xia, Menghan and Fu, Xiao and Wang, Xintao and Mu, Lianrui and Cao, Jinwen and Liu, Zuozhu and Hu, Haoji and Bai, Xiang and Wan, Pengfei and others},
  journal={arXiv preprint arXiv:2503.11647},
  year={2025}
}

@article{dfot,
  title={History-Guided Video Diffusion},
  author={Song, Kiwhan and Chen, Boyuan and Simchowitz, Max and Du, Yilun and Tedrake, Russ and Sitzmann, Vincent},
  journal={arXiv preprint arXiv:2502.06764},
  year={2025}
}

@article{t2icomp,
  title={T2i-compbench: A comprehensive benchmark for open-world compositional text-to-image generation},
  author={Huang, Kaiyi and Sun, Kaiyue and Xie, Enze and Li, Zhenguo and Liu, Xihui},
  journal={Advances in Neural Information Processing Systems},
  volume={36},
  pages={78723--78747},
  year={2023}
}

@article{t2vcomp,
  title={T2v-compbench: A comprehensive benchmark for compositional text-to-video generation},
  author={Sun, Kaiyue and Huang, Kaiyi and Liu, Xian and Wu, Yue and Xu, Zihan and Li, Zhenguo and Liu, Xihui},
  journal={arXiv preprint arXiv:2407.14505},
  year={2024}
}

@inproceedings{controlnet,
  title={Adding conditional control to text-to-image diffusion models},
  author={Zhang, Lvmin and Rao, Anyi and Agrawala, Maneesh},
  booktitle={Proceedings of the IEEE/CVF international conference on computer vision},
  pages={3836--3847},
  year={2023}
}

@inproceedings{t2iadapter,
  title={T2i-adapter: Learning adapters to dig out more controllable ability for text-to-image diffusion models},
  author={Mou, Chong and Wang, Xintao and Xie, Liangbin and Wu, Yanze and Zhang, Jian and Qi, Zhongang and Shan, Ying},
  booktitle={Proceedings of the AAAI conference on artificial intelligence},
  year={2024}
}

@inproceedings{animate-anyone,
  title={Animate anyone: Consistent and controllable image-to-video synthesis for character animation},
  author={Hu, Li},
  booktitle={Proceedings of the IEEE/CVF Conference on Computer Vision and Pattern Recognition},
  pages={8153--8163},
  year={2024}
}

@article{fulldit,
  title={Fulldit: Multi-task video generative foundation model with full attention},
  author={Ju, Xuan and Ye, Weicai and Liu, Quande and Wang, Qiulin and Wang, Xintao and Wan, Pengfei and Zhang, Di and Gai, Kun and Xu, Qiang},
  journal={arXiv preprint arXiv:2503.19907},
  year={2025}
}

@article{drag-a-video,
  title={Drag-a-video: Non-rigid video editing with point-based interaction},
  author={Teng, Yao and Xie, Enze and Wu, Yue and Han, Haoyu and Li, Zhenguo and Liu, Xihui},
  journal={arXiv preprint arXiv:2312.02936},
  year={2023}
}

@article{control-a-video,
  title={Control-a-video: Controllable text-to-video generation with diffusion models},
  author={Chen, Weifeng and Ji, Yatai and Wu, Jie and Wu, Hefeng and Xie, Pan and Li, Jiashi and Xia, Xin and Xiao, Xuefeng and Lin, Liang},
  journal={arXiv e-prints},
  pages={arXiv--2305},
  year={2023}
}

@article{controlnext,
  title={Controlnext: Powerful and efficient control for image and video generation},
  author={Peng, Bohao and Wang, Jian and Zhang, Yuechen and Li, Wenbo and Yang, Ming-Chang and Jia, Jiaya},
  journal={arXiv preprint arXiv:2408.06070},
  year={2024}
}

@article{ar-diffusion,
  title={Ar-diffusion: Asynchronous video generation with auto-regressive diffusion},
  author={Sun, Mingzhen and Wang, Weining and Li, Gen and Liu, Jiawei and Sun, Jiahui and Feng, Wanquan and Lao, Shanshan and Zhou, SiYu and He, Qian and Liu, Jing},
  journal={arXiv preprint arXiv:2503.07418},
  year={2025}
}

@article{causalfusion,
  title={Causal diffusion transformers for generative modeling},
  author={Deng, Chaorui and Zhu, Deyao and Li, Kunchang and Guang, Shi and Fan, Haoqi},
  journal={arXiv preprint arXiv:2412.12095},
  year={2024}
}

@inproceedings{physgen,
  title={Physgen: Rigid-body physics-grounded image-to-video generation},
  author={Liu, Shaowei and Ren, Zhongzheng and Gupta, Saurabh and Wang, Shenlong},
  booktitle={European Conference on Computer Vision},
  pages={360--378},
  year={2024},
  organization={Springer}
}

@article{diffusion-as-shader,
  title={Diffusion as Shader: 3D-aware Video Diffusion for Versatile Video Generation Control},
  author={Gu, Zekai and Yan, Rui and Lu, Jiahao and Li, Peng and Dou, Zhiyang and Si, Chenyang and Dong, Zhen and Liu, Qifeng and Lin, Cheng and Liu, Ziwei and others},
  journal={arXiv preprint arXiv:2501.03847},
  year={2025}
}

@article{worldmodelbench,
  title={Worldmodelbench: Judging video generation models as world models},
  author={Li, Dacheng and Fang, Yunhao and Chen, Yukang and Yang, Shuo and Cao, Shiyi and Wong, Justin and Luo, Michael and Wang, Xiaolong and Yin, Hongxu and Gonzalez, Joseph E and others},
  journal={arXiv preprint arXiv:2502.20694},
  year={2025}
}

@article{far,
  title={Long-Context Autoregressive Video Modeling with Next-Frame Prediction},
  author={Gu, Yuchao and Mao, Weijia and Shou, Mike Zheng},
  journal={arXiv preprint arXiv:2503.19325},
  year={2025}
}

@article{gen3c,
  title={Gen3c: 3d-informed world-consistent video generation with precise camera control},
  author={Ren, Xuanchi and Shen, Tianchang and Huang, Jiahui and Ling, Huan and Lu, Yifan and Nimier-David, Merlin and M{\"u}ller, Thomas and Keller, Alexander and Fidler, Sanja and Gao, Jun},
  journal={arXiv preprint arXiv:2503.03751},
  year={2025}
}

@inproceedings{dmd,
  title={One-step diffusion with distribution matching distillation},
  author={Yin, Tianwei and Gharbi, Micha{\"e}l and Zhang, Richard and Shechtman, Eli and Durand, Fredo and Freeman, William T and Park, Taesung},
  booktitle={Proceedings of the IEEE/CVF conference on computer vision and pattern recognition},
  pages={6613--6623},
  year={2024}
}

@article{deepseek-r1,
  title={Deepseek-r1: Incentivizing reasoning capability in llms via reinforcement learning},
  author={Guo, Daya and Yang, Dejian and Zhang, Haowei and Song, Junxiao and Zhang, Ruoyu and Xu, Runxin and Zhu, Qihao and Ma, Shirong and Wang, Peiyi and Bi, Xiao and others},
  journal={arXiv preprint arXiv:2501.12948},
  year={2025}
}

@article{qwen,
  title={Qwen technical report},
  author={Bai, Jinze and Bai, Shuai and Chu, Yunfei and Cui, Zeyu and Dang, Kai and Deng, Xiaodong and Fan, Yang and Ge, Wenbin and Han, Yu and Huang, Fei and others},
  journal={arXiv preprint arXiv:2309.16609},
  year={2023}
}

@article{self-forcing,
  title={Self Forcing: Bridging the Train-Test Gap in Autoregressive Video Diffusion},
  author={Huang, Xun and Li, Zhengqi and He, Guande and Zhou, Mingyuan and Shechtman, Eli},
  journal={arXiv preprint arXiv:2506.08009},
  year={2025}
}

@article{longlive,
  title={LongLive: Real-time Interactive Long Video Generation},
  author={Yang, Shuai and Huang, Wei and Chu, Ruihang and Xiao, Yicheng and Zhao, Yuyang and Wang, Xianbang and Li, Muyang and Xie, Enze and Chen, Yingcong and Lu, Yao and Han, Song and Chen, Yukang},
  journal={arXiv preprint arXiv:2509.22622},
  year={2025}
}

@article{rolling-forcing,
  title={Rolling Forcing: Autoregressive Long Video Diffusion in Real Time},
  author={Liu, Kunhao and Hu, Wenbo and Xu, Jiale and Shan, Ying and Lu, Shijian},
  journal={arXiv preprint arXiv:2509.25161},
  year={2025}
}

@article{svi,
  title={Stable Video Infinity: Infinite-Length Video Generation with Error Recycling},
  author={Li, Wuyang and Pan, Wentao and Luan, Po-Chien and Gao, Yang and Alahi, Alexandre},
  journal={arXiv preprint arXiv:2510.09212},
  year={2025}
}

@article{multigen,
  title={MultiGen: Level-Design for Editable Multiplayer Worlds in Diffusion Game Engines},
  author={Po, Ryan and Zhang, David Junhao and Hertz, Amir and Wetzstein, Gordon and Wadhwa, Neal and Ruiz, Nataniel},
  journal={arXiv preprint arXiv:2603.06679},
  year={2026}
}

@article{multiworld,
  title={MultiWorld: Scalable Multi-Agent Multi-View Video World Models},
  author={Wu, Haoyu and Yu, Jiwen and Zou, Yingtian and Liu, Xihui},
  journal={arXiv preprint arXiv:2604.18564},
  year={2026}
}

@article{solaris,
  title={Solaris: Building a Multiplayer Video World Model in Minecraft},
  author={Savva, Georgy and Michel, Oscar and Lu, Daohan and Waiwitlikhit, Suppakit and Meehan, Timothy and Mishra, Dhairya and Poddar, Srivats and Lu, Jack and Xie, Saining},
  journal={arXiv preprint arXiv:2602.22208},
  year={2026}
}

@article{3dgs,
  title={3D Gaussian Splatting for Real-Time Radiance Field Rendering},
  author={Kerbl, Bernhard and Kopanas, Georgios and Leimk{\"u}hler, Thomas and Drettakis, George},
  journal={ACM Transactions on Graphics},
  volume={42},
  number={4},
  year={2023}
}

\newpage
\begin{appendices}

\noindent\textit{Scope note.} This is a position paper. Its contribution lies in systematically arguing for a research direction and stimulating productive community debate, not in demonstrating that the proposed technology is already mature. We treat the feasibility of this direction as a reasoned research hypothesis supported by progress on individual capabilities, rather than an established result for their integration into a complete engine. The alternative views in Sec.~\ref{sec:alternative} and Appendix~\ref{appendix:alternative} are included precisely because legitimate disagreement exists on these questions, and engaging with that disagreement head-on is central to the paper's purpose.

Each appendix focuses on a specific aspect of IGV-based game engines (alternative views, ethical issues, maturity criteria, an illustrative workflow) and is designed to support the main argument rather than stand alone as a comprehensive treatment of its topic. The main text (Secs.~1--9) carries the core position; the appendices provide supporting context at a level appropriate for a position paper. Deeper treatment of individual topics (e.g., multi-agent netcode, designer tooling, emergence metrics) is left to follow-up work and community discussion. Discussions herein, including alternative views and their responses, draw on our own research experience and extensive interactions with researchers, game developers, and community participants across multiple venues.

\section{Preliminaries}
\label{appendix:preliminary}
\subsection{Video Generation Models}

\textbf{Video Generation Models} aim to synthesize temporally consistent video sequences \( \textbf{x} = \{x_0, x_1, \dots, x_t\} \), where $x_i$ indicates the $i$-th frame. The video is generated by an autoregressive model~\cite{razavi2019generating,videogpt,videopoet,emu3}, a diffusion model~\cite{imagen,sora}, or a masked transformer model~\cite{maskgit,magvit}. With the rise of diffusion models~\cite{ddpm, sde, score,flow,rectified}, which have now become the mainstream approach in video generation due to its high-quality generation performance, significant progress has been achieved in this field~\cite{sora,cogvideox,kling,veo2,seaweed,moviegen,runway,luma,vidu,wan,hunyuanvideo}.

\textbf{Conditional Video Generation} is formulated as \( p(\textbf{x}|c) \) and it varies depending on the type of control signal \( c \) which denotes conditions such as text prompts or other control signal. 
Approaches such as~\cite{guo2025sparsectrl, ni2023conditional, dynamicrafter} incorporate images as control signals for the video generator, improving both video quality and temporal relationship modeling.
Methods such as Direct-a-Video~\cite{direct-a-video}, MotionCtrl~\cite{motionctrl}, and CameraCtrl~\cite{cameractrl} use camera embedders to adjust camera poses, enabling control over camera movements in generated videos. 3DTrajMaster~\cite{3dtrajmaster} extends this capability by transforming 2D camera signals into 3D for more advanced control. ReCamMaster~\cite{recammaster} re-shoots the source video with novel camera trajectories.

\textbf{Autoregressive Video Generation Models.}
Since game videos require variable-length or even infinite-length generation to enable interactive game experience, autoregressive mechanism is necessary. 
Autoregressive video generation refers to a process where new frames are generated based on previously generated frames, which can be expressed as $p(x_1,x_2,...,x_n)=\prod_{i=1}^np(x_i|x_1,x_2,...,x_{i-1})$, where $x_i$ indicates the $i$-th frame.
An intuitive approach is to adopt GPT-like next-token prediction methods~\cite{videopoet,nova,emu3}; however, this approach often falls short in terms of generation quality.
Relying on the exceptional performance of diffusion models, Diffusion Forcing~\cite{cdf, dfot} implements autoregression by applying different levels of noise to different frames, allowing the denoising of new frames with higher noise levels while conditioning on previous frames with lower noise levels. Methods~\cite{gamefactory,matrix,gamengine,oasis} leveraging Diffusion Forcing has achieved remarkable results.

\subsection{AI-driven Game Applications}
\textbf{Game Video Generation.} Previous works have utilized GANs~\cite{gamegan,drivegan,caddy} to generate game videos or used NeRF to reconstruct 3D scenes to simulate the game process~\cite{pe,pgm}, but often fall short in terms of generation quality. 
Using the powerful generative capabilities of diffusion models, some works~\cite{diamond,gamengine,playgen,oasis,matrix} have produced high-quality game videos. However, the content is typically confined to specific preexisting games.
Open-domain Methods~\cite{genie2,matrix,gamefactory}, by utilizing multi-stage training or large-scale training datasets, create new content for video games.

\textbf{Game Design Assistant.} AI-powered design assistants offer numerous advantages throughout the creative process, depending on the tool, the type of AI and the creative workflow. These systems can streamline development, reduce costs, reduce manual effort, improve team collaboration, and even inspire creativity~\cite{liapis2015searching}. In the gaming domain, most existing AI-driven design tools primarily assist by auto-completing an ongoing design~\cite{tanagra} or generating multiple design suggestions for creators to evaluate~\cite{sketchbook,susketch,baba,lottery}.

\textbf{Intelligent Game Agent.}
Reinforcement learning has long been the predominant approach in this domain. Early efforts explored the use of hierarchical RL~\cite{gsb,clip4mc,minedojo,juewu,zerorl,minedreamer,plan4mc} in the context of MineRL competitions~\cite{minerl}. However, due to the absence of guidance from prior knowledge, such approaches often struggle to perform effectively on long-horizon tasks.
With the advancement of LLM~\cite{gpt4,llama}, leveraging their prior knowledge to plan long-horizon tasks has shown promising results. Recent advancements in LLM-related research~\cite{voyager,deps,gitm,mp5,octopus,llmplanner} have significantly propelled the progress of agents in long-horizon tasks.

\newpage
\section{Overview Table for Levels of GGE}
\label{appendix:levels}
In Table~\ref{tab:levels}, we demonstrate the overview of different maturity levels for GGE.
\begin{table*}[h]
    \centering
    \renewcommand{\arraystretch}{1.5} % 调整行高
    \renewcommand{\cellalign}{vh} % 单元格内容垂直和水平居中
    \begin{tabular}{|c|>{\centering\arraybackslash}p{2cm}|>{\centering\arraybackslash}p{3.48cm}|>{\centering\arraybackslash}p{3.48cm}|>{\centering\arraybackslash}p{2cm}|}
    \hline
    \textbf{Level} & \textbf{Name} & \textbf{Technical Features} & \textbf{Application Examples} & \textbf{Category}  \\
    \hline
    L0 & No AI-Assisted Assets Generation & Manual creation and integration of all game assets and logic. & \textit{Super Mario}: fixed levels; \textit{Tetris}: fixed rules. & \multirow{2}{=}{Traditional Manual Game Development} \\
    \cline{1-4}
    L1 & AI-Assisted Assets Generation & AI-assisted
creation and integration of
game assets and logic. & \textit{Cyberpunk 2077}: AI-generated assets; \textit{AI Dungeon}: real-time NPC dialogues. & \\
    \hline
    L2 & Physics-Compliant Interactive World Generation & Real-time \textbf{physics-compliant} video generation with user \textbf{interactions}, supported by the Dynamics module. & E.g., Player sets fire to wooden bridges, AI dynamically renders blazing spans and rerouted enemy paths & \multirow{3}{=}{Next-Gen AI-Driven Generative Game Engine} \\
    \cline{1-4}
    L3 & Causal-Reasoning World Simulation & World simulation with \textbf{causal reasoning} across time based on L2, incorporating the Intelligence module. & E.g., Killing a faction leader in Act 1 triggers city-wide riots in Act 3. & \\
    \cline{1-4}
    L4 & Self-Evolving World Ecosystem & Autonomous world \textbf{evolution} with \textbf{emergent} behaviors based on L2 and L3, requiring advanced Intelligence module. & E.g. NPCs self-organize governments and trade as population increases. & \\
    \hline
    \end{tabular}
    \caption{Proposed Maturity Levels (L0-L4) of Generative Game Engine. L0-L1 represent traditional manual game development with limited AI assistance, while L2-L4 showcase next-generation game engines featuring video-based world generation.}
    \vspace{-0.5cm}
    \label{tab:levels}
    \end{table*}

\newpage
\section{Observable Criteria for GGE Maturity Levels}
\label{appendix:criteria}

These proposed criteria make the L0--L4 roadmap (Sec.~\ref{sec:levels}) actionable. They are \emph{necessary but not sufficient}: severe failures may disqualify a system. In principle, each level should retain earlier capabilities, although transitional systems may fall short. Numerical thresholds (e.g., $<$150ms latency for L2) are provisional 2026 targets and should be recalibrated as hardware and algorithms evolve.

\subsection{L0: No AI-Assisted Assets Generation}
\begin{itemize}
    \item Does the system use any AI generative model in any production stage? (No $\rightarrow$ L0)
    \item Are all game assets fully fixed before release, with no runtime dynamic generation? (Yes $\rightarrow$ L0)
\end{itemize}

\subsection{L1: AI-Assisted Assets Generation}
\begin{itemize}
    \item Does AI participate in \emph{design-time} content generation (textures, dialogue, concept art, etc.)? (Yes)
    \item Is generated content still subject to human curation before being baked into the game? (Yes $\rightarrow$ L1)
    \item Runtime AI generation, if present, does not substantively affect gameplay. If it does, evaluate against L2 criteria.
\end{itemize}

\subsection{L2: Physics-Compliant Interactive World Generation}
\begin{itemize}
    \item Can the system generate physically plausible video output for \emph{interactions not seen in training data} (e.g., ``slicing a pumpkin with a sword'')?
    \item Does the generated video's physical consistency (gravity direction, collision response, etc.) meet a threshold on established benchmarks such as WorldSimBench~\cite{worldsimbench}?
    \item Is the end-to-end latency from user input to visual feedback below the interactive gaming threshold? (Suggested: \finalrev{$<$150ms} for L2)
    \item \newrevised{For multiplayer-capable systems: is cross-player world state consistency maintained under concurrent inputs, with synchronization latency below interactive thresholds? (Suggested: $<$50ms for competitive genres)}
\end{itemize}

\subsection{L3: Causal-Reasoning World Simulation}
\begin{itemize}
    \item Can the system maintain causal consistency of world state over \emph{extended gameplay sessions} ($>$10 minutes)? (e.g., an NPC killed at minute 1 should not reappear at minute 11)
    \item Given an initial event, can the system generate \emph{plausible long-horizon consequences} (dozens of steps ahead), beyond short-term extrapolation?
    \item Can the system handle \emph{non-local causality}? (e.g., a distant event indirectly affecting the current scene)
\end{itemize}

\subsection{L4: Self-Evolving World Ecosystem}
Defining operational criteria for L4 is itself an open research problem, as ``emergence'' resists clean formalization. The following are preliminary indicators rather than definitive tests; refining them is an important direction for future work. Self-evolution remains bounded by designer-specified rules and style constraints, with interventions available to preserve each game's authored identity.
\begin{itemize}
    \item In \emph{unattended} operation, does the virtual world exhibit behavior patterns, social structures, or rules that are \emph{not explicitly present in training data}?
    \item Are these emergent behaviors \emph{self-consistent} over extended periods (no collapse, no degenerate loops, no obvious contradictions)?
    \item Are emergent rules \emph{observable and analyzable} by researchers, rather than purely black-box? Existing multi-agent studies (e.g., CAMEL~\cite{camel}) provide only weak proxies for world-level emergence; stronger evaluation protocols remain to be developed.
\end{itemize}

\newpage
\section{Alternative Views: Detailed Analysis}
\label{appendix:alternative}

Below we provide detailed responses to each alternative view raised in Sec.~\ref{sec:alternative}.

\subsection{Alternative View \#1: GGE vs.\ Procedural Content Generation}

\begin{tcolorbox}[colframe=gray,colback=gray,coltext=white,boxsep=0.5mm,before skip=3pt,after skip=3pt]
While GGE represents an automated approach to game content generation, it is worth examining whether it shares the same limitations as Procedural Content Generation (PCG) methods, specifically the tendency to produce repetitive content and the presence of difficult-to-fix bugs that potentially limit their practical applications.
\end{tcolorbox}

GGE aims to extend the diversity of procedural content generation by learning content and world priors from large datasets, rather than relying only on fixed asset combinations. AI image generation and its use on Civitai~\cite{civitai} illustrate the potential for diverse outputs. This distinction does not guarantee meaningful variety or eliminate bugs: rule verification, controllability, and debugging tools remain necessary, as does designer judgment about whether generated content serves the game.

\subsection{Alternative View \#2: Traditional Rendering vs.\ IGV}

\begin{tcolorbox}[colframe=gray,colback=gray,coltext=white,boxsep=0.5mm,before skip=3pt,after skip=3pt]
Given that traditional rendering pipelines in standard game engines offer efficient asset rendering and allow more resources to be allocated to gameplay enhancement, why should we adopt IGV instead of maintaining the traditional approach that prioritizes gameplay dynamics over graphical realism?
\end{tcolorbox}

Traditional rendering pipelines efficiently handle graphics, allowing developers to focus more resources on gameplay rather than graphics. This raises concerns that IGV might shift too many resources toward visual realism at the expense of gameplay quality, a trade-off that many developers would find unreasonable.
However, IGV represents a paradigm shift that enhances both graphical realism and gameplay quality together, rather than trading off one for the other. We analyze this from four aspects:
(1) \textbf{Complete system}: IGV is not just a generation model, but a system integrating Control, Memory, Dynamics and Intelligence. Beyond improving graphics, it enhances gameplay through, for example, personalized game content, infinite explorable experiences, and intelligent NPC behaviors.
(2) \textbf{Enhanced gaming experience}: IGV enables dynamic, customized, and infinitely explorable experiences, which traditional game development with triangle rendering and added gameplay logic cannot easily achieve.
(3) \textbf{Enhanced creative freedom}: IGV's virtual world generation capabilities free developers from focusing on graphics, allowing them to concentrate on gameplay design. Its controllability enables developers to freely exercise creativity in designing more innovative gameplay experiences.
(4) \textbf{Positive industry impact}: IGV's efficiency and capabilities accelerate game development and lower entry barriers. This attracts more developers, resulting in more creative games and enriching the gaming experience industry-wide.

\subsection{Alternative View \#3: Economic Costs}

\begin{tcolorbox}[colframe=gray,colback=gray,coltext=white,boxsep=0.5mm,before skip=3pt,after skip=3pt]
The economic costs of GGE appear to be significant. For instance, the computational overhead is substantial since GGE relies on IGV and LLMs, which are computationally intensive large models. Do these costs prevent IGV-centered GGE from becoming the next generation of game engines? Are these costs we incur for implementing GGE justified by the benefits it brings?
\end{tcolorbox}

Development savings and runtime affordability are separate questions. Generating assets or interactions may reduce some authoring work, but training, deployment, and continuous inference add costs. In the near term, players may need cloud services or higher-end local hardware, paying through service charges, hardware purchases, electricity use, and waiting time. Development savings do not establish lower total cost or lower prices for players.

Research offers concrete avenues for reducing these costs: CausVid~\cite{causvid} uses distribution matching distillation (DMD)~\cite{dmd}, while Cosmos~\cite{cosmos} explores efficient generation. Distillation, speculative decoding, key-value caching, quantization, parallel execution, and lower-resolution generation can reduce inference demands in suitable settings. Memory reuse may also avoid redundant generation. These techniques motivate continued research, but do not guarantee affordable, low-latency generation with the visual quality, consistency, and control required by a particular game.

Beyond computational costs, other economic considerations include:
\begin{itemize}
    \item \textbf{Data Collection Costs}: Shared datasets such as GameGen-X~\cite{gamegenx} may reduce duplicated collection effort, but game-specific data, annotation, and model adaptation remain costs to account for.
    \item \textbf{Licensing Costs}: Rights to training data and generated content require appropriate arrangements; creating new IP does not remove these obligations or their costs.
    \item \textbf{Safety Control Costs}: Moderation, testing, and human oversight add development and runtime costs. Commercial generative products such as Runway~\cite{runway}, Midjourney~\cite{midjourney}, and ChatGPT~\cite{chatgpt} motivate exploring viable deployments, but do not demonstrate that benefits outweigh costs for IGV games.
\end{itemize}

Progress in LLM capabilities and accessibility, illustrated by DeepSeek-R1~\cite{deepseek-r1}, GPT-4o~\cite{gpt-4o-image-generation}, and Qwen~\cite{qwen}, provides an analogy for improvement, not an economic forecast for video generation. The relevant comparison must include total development costs and recurring costs over actual play time, alongside the value players place on the resulting experience.

Our position therefore does not require replacing every traditional deployment. Hybrid pipelines, selective generation, and traditional rendering remain appropriate where they offer better cost or latency, particularly on low-power devices and in precision-critical games. IGV is worth pursuing where its potential for responsive worlds and personalized content can justify the total cost; whether that trade-off succeeds remains dependent on the game and deployment.

\subsection{Alternative View \#4: Video Generation vs.\ 3D Scene Generation}

\begin{tcolorbox}[colframe=gray,colback=gray,coltext=white,boxsep=0.5mm,before skip=3pt,after skip=3pt]
3D scene generation methods offer explicit geometry that naturally integrates with existing engine pipelines. Why should video generation, rather than 3D generation, serve as the core of next-generation game engines?
\end{tcolorbox}

We acknowledge that 3D scene generation (e.g., 3D Gaussian Splatting~\cite{3dgs}, point cloud reconstruction) offers clear advantages in geometric precision and compatibility with existing engine pipelines that rely on explicit 3D representations. However, we argue that video generation holds fundamental advantages as the \emph{core} modality for several reasons:

(1) \textbf{Data scale and accessibility}: Video data is vastly more abundant and diverse than 3D data. The internet provides an essentially unlimited supply of videos capturing real-world dynamics, appearances, and interactions, whereas large-scale, high-quality 3D datasets remain scarce and expensive to produce. This data advantage directly translates to stronger generative priors.

(2) \textbf{Unified expressiveness}: Video naturally encodes appearance, dynamics, physics, and interactions in a single modality. In contrast, 3D representations typically capture geometry and appearance but require separate systems for animation, physics, and logic---reintroducing the complexity of traditional engines.

(3) \textbf{Complementary rather than competing}: We do not claim that 3D representations are unnecessary. In fact, 3D structures can serve as the memory and spatial backbone within an IGV-based GGE. For example, reconstructing 3D scene representations from generated videos~\cite{see3d,viewcrafter,vid2sim} can enhance spatial consistency and enable save/load functionality. The position is that \emph{video generation drives the core loop}, while 3D representations play a supporting role in the system architecture.

This is compatible with hybrid world state: frame history preserves visual detail, latent memory supports compact context, explicit spatial structures constrain geometry, and symbolic variables encode game rules. Developer tools should expose ways to query, edit, inspect, and restore this state. Keeping these representations synchronized remains an open challenge.

\subsection{Alternative View \#5: Traditional Engines + LLM vs.\ IGV Paradigm Shift}

\begin{tcolorbox}[colframe=gray,colback=gray,coltext=white,boxsep=0.5mm,before skip=3pt,after skip=3pt]
Augmenting existing engines like Unreal or Unity with LLM-driven code and asset generation could incrementally achieve similar goals. Why is a paradigm shift to IGV necessary?
\end{tcolorbox}

The incremental path---using LLMs to generate code, scripts, and assets within traditional engines---is a reasonable near-term strategy, and we expect it to coexist with IGV-based approaches. However, this path faces fundamental ceilings that IGV can overcome:

(1) \textbf{Combinatorial limits of pre-made assets}: Traditional engines, even augmented with LLM-generated code, ultimately compose games from discrete, pre-defined asset types (meshes, textures, animations, scripts). The expressiveness of such compositions is bounded by the asset vocabulary. IGV operates at the \emph{perception level}, generating continuous visual experiences that are not constrained by a fixed asset inventory.

(2) \textbf{Explicit rule brittleness}: Traditional engines require explicit logic rules for physics, NPC behavior, and game mechanics. LLMs can help write these rules, but the fundamental paradigm of rule-based logic remains---with its inherent limitations in handling emergent, unscripted situations. IGV learns implicit world rules from data, enabling more natural and diverse in-game behaviors.

(3) \textbf{Long-term coexistence}: We acknowledge that both paths will coexist for a considerable period. The incremental path excels where precision and determinism are paramount (e.g., competitive multiplayer games), while IGV excels where open-ended content and creative freedom are prioritized. Our position is that IGV offers the higher long-term ceiling---not because the incremental path is without value, but because the limitations are structural to engines organized around discrete asset composition and explicit rule logic.

\subsection{Alternative View \#6: Capability Advantages \texorpdfstring{$\neq$}{do not imply} Good Game Engine}

\begin{tcolorbox}[colframe=gray,colback=gray,coltext=white,boxsep=0.5mm,before skip=3pt,after skip=3pt]
Strong generalization, physical realism, and controllability do not automatically translate into a good game engine or good games. How do IGV's technical strengths connect to actual game quality?
\end{tcolorbox}

This is an important logical distinction that we take seriously. We clarify our position as follows:

(1) \textbf{Engine capability vs.\ game quality}: A game engine is a \emph{tool}; a good game is the product of humans using that tool creatively. We do not claim that IGV alone produces good games. Rather, we argue that IGV \emph{expands the capability boundary} of what game engines can offer, enabling new categories of experiences that are infeasible with current engines---such as truly infinite open worlds, fully personalized content, and dynamically generated narratives.

(2) \textbf{Capability boundaries determine product possibilities}: While a minimalist pixel game can be deeply engaging, this does not diminish the value of expanding engine capabilities. By analogy, Unreal Engine's Nanite technology does not guarantee better games, but it enables visual fidelity that was previously impossible, opening new creative possibilities. Similarly, IGV's generative capabilities open possibilities that fixed-asset engines cannot provide.

(3) \textbf{Residual uncertainty}: We acknowledge that the gap between ``IGV can generate impressive content'' and ``IGV enables commercially successful games'' remains to be bridged. This requires advances not only in generation quality but also in controllability, consistency, and developer tooling. Our maturity roadmap (L0--L4) explicitly charts this progression, and we do not claim that the endpoint has been reached.

\subsection{Alternative View \#7: Does IGV Need 100\% Reliability?}

\begin{tcolorbox}[colframe=gray,colback=gray,coltext=white,boxsep=0.5mm,before skip=3pt,after skip=3pt]
IGV cannot match traditional engines in physical accuracy or control precision. If it cannot achieve 100\% reliability in physics simulation and player control, why should it replace proven, deterministic systems?
\end{tcolorbox}

This challenge conflates two sub-questions that share the same ``not 100\% $\Rightarrow$ not useful'' attack pattern. We address both and then honestly state the residual uncertainty.

\textbf{Physics accuracy: coverage vs.\ precision.}
We do not claim IGV surpasses traditional engines on formula-based physics (rigid body mechanics, Newtonian dynamics). For these phenomena, explicit solvers remain superior and can serve as external control signals within IGV (as discussed in the Dynamics module, Sec.~\ref{subsec:dynamics}). Our claim is that IGV and traditional physics engines are \emph{complementary}: traditional solvers handle formula-based physics, while IGV handles \emph{formula-hard} physics (fluid deformation, cloth wrinkling, fire propagation, soft-body collisions) where writing explicit equations is itself extremely difficult, and \emph{semantic} physics (``this bottle should shatter'', ``this character should flinch'') that lies entirely beyond traditional engine scope. Evaluating IGV by rigid-body precision alone is the wrong benchmark, analogous to dismissing deep learning because SIFT features are more precise on a narrow task.

Moreover, games have never required 100\% physical correctness. \textit{The Legend of Zelda: Breath of the Wild}'s chemistry system is far from real physics, yet its designable, tunable simplified physics created iconic gameplay. Physics Tuning (Sec.~\ref{subsec:dynamics}) is in fact an advantage of IGV over pure simulators: IGV can both learn real-world physics \emph{and} deviate from reality to serve gameplay design.

\textbf{Control: dimension expansion, not precision competition.}
The concern that IGV merely replicates WASD control is a dimension mismatch. IGV's control value lies in \emph{expanding} what is controllable, not competing on existing controls:
(1) \textbf{Semantic granularity}: from discrete key-to-animation mappings to intent-driven continuous behavior generation (e.g., ``slam the door angrily'' vs.\ ``close the door gently'').
(2) \textbf{Long-tail action coverage}: traditional engines must pre-build assets for every action; long-tail actions (e.g., ``clumsily climb over a fence'') are cut due to cost. IGV learns long-tail action distributions from video data.
(3) \textbf{Compositionality}: ``swimming + injured + holding a torch'' requires multiplicative animation costs in traditional engines; IGV's generative compositionality handles such combinations naturally.
The correct question is not ``can IGV do WASD'' (yes), but ``can traditional engines do what IGV uniquely offers'' (no, or at prohibitive cost).

\textbf{Residual uncertainty.}
We acknowledge two risks this view raises that we cannot fully resolve: (1) \emph{Runtime stability}: unlike design-time generation where creators can cherry-pick results, runtime generation quality directly affects player experience. Current IGV models may still produce degraded outputs during long sessions. This gap must be crossed between L2 and L3 on our roadmap. (2) \emph{Determinism for competitive games}: esports and speedrun scenarios demand frame-level determinism that pure generative approaches may not guarantee. IGV may need hybrid deployment with traditional pipelines in these niches. We do not shy away from these uncertainties; identifying \emph{which} uncertainties are worth the community's investment is precisely the role of a position paper.

\newrevised{\subsection{Alternative View \#8: Hybrid Multi-Technique Composition vs.\ IGV-as-Core}

\begin{tcolorbox}[colframe=gray,colback=gray,coltext=white,boxsep=0.5mm,before skip=3pt,after skip=3pt]
Different game components (3D assets, interaction scripts, NPC behavior) could each adopt the best-suited generative technique. Why insist on IGV as a single unifying backbone rather than letting each game type mix and match optimal technologies?
\end{tcolorbox}

This is a pragmatic and reasonable view that we partly embrace. Our response clarifies what ``IGV-as-core'' does and does not claim:

(1) \textbf{A unified perceptual layer is still needed.} Even in a heterogeneous pipeline, something must integrate the outputs of separate modules (3D assets, dialogue, physics) into a coherent visual experience. IGV provides this integration layer: it renders the combined world state into continuous video output, aiming to maintain visual coherence across modules. Agreement between their states and the generated observations remains an open integration requirement.

(2) \textbf{Our framework already supports heterogeneity.} The six-module GGE architecture explicitly accommodates specialized sub-systems: the Dynamics module can invoke traditional physics solvers, the Intelligence module can leverage LLMs, and the GamePlay module can use RL-based agents. Calling IGV the ``core'' does not mean it handles every sub-task alone; it means IGV provides the foundational world representation on which other modules operate.

(3) \textbf{The long-term convergence argument.} As unified multimodal models improve (Sec.~\ref{subsec:generation}), the boundaries between separately-trained specialist modules are likely to blur. A single video-native world model that internalizes physics, control, and reasoning end-to-end may ultimately outperform loosely-coupled pipelines, analogous to how end-to-end learned systems have overtaken modular pipelines in autonomous driving perception.

\finalrev{We also acknowledge that our position is strongest for games where visual perception and dynamic world generation are central (open-world, sandbox, narrative adventures); for genres with minimal world dynamics (card games, text adventures), IGV's role as core is naturally weaker.} We acknowledge that in the near term, hybrid composition will dominate practical deployments. Our position is about the long-term architectural direction, not a prescription for today's production pipelines.}

\subsection{Alternative View \#9: Why Would Players Want an IGV Game?}
\label{appendix:playerexperience}

\begin{tcolorbox}[colframe=gray,colback=gray,coltext=white,boxsep=0.5mm,before skip=3pt,after skip=3pt]
More freedom and more content need not make a game enjoyable. Players may want a particular authored game, rather than an endlessly changing generator. What incentive would they have to choose IGV?
\end{tcolorbox}

The potential incentive is meaningful responsiveness: worlds that react to unusual player actions, exploration beyond a fixed content inventory, or narratives and challenges adapted to a player's choices. These are possibilities enabled by the engine, not evidence that unlimited content is intrinsically desirable. Repetition, filler, poor pacing, or unpredictable mechanics can undermine the experience regardless of how content is produced.

Our claim is that IGV should provide tools under designer control. Designers specify the game's goals, rules, art direction, pacing, and acceptable variation through control signals, constraints, and intervention. A player choosing ``game X'' should still encounter its intended identity; neither generative content nor L4 self-evolution requires different games to converge to the same experience. Deliberately stylized or non-physical settings can remain part of that design.

Good games still depend on how designers use these capabilities, and players may reasonably prefer a finite, tightly authored experience. IGV must earn its place through the resulting play experience, taking into account the money, hardware, energy, and latency costs discussed in Alternative View \#3. Our position motivates expanding the available tools; it does not guarantee enjoyment, adoption, or commercial success for any title.

\newpage
\section{Ethical Issues}
\label{appendix:ethical}

\begin{tcolorbox}[colframe=gray,colback=gray,coltext=white]
\textbf{Copyright Issues}: How should we determine copyright ownership and protect legitimate copyright interests of all parties involved in GGE-generated games?
\end{tcolorbox}

Copyright protection presents a new challenge in generative AI development, including IGV, which requires significant attention from both technical and legal perspectives. While this is a complex issue, the industry is actively working towards solutions that can foster the mutual development of AI technology and copyright protection. To address these challenges, we propose several approaches:

Training data for IGV should prioritize the use of copyright-free or properly licensed data sources to minimize legal risks.
Game developers can build mutually beneficial partnerships with copyright holders to legally obtain data and share the copyright of the created content. For instance, while noting the Studio Ghibli's recent copyright concerns with OpenAI, we observe that Ghibli has successful experiences in collaborating with game companies (e.g., development of game "Ni no Kuni"). Such examples demonstrate the feasibility and value of proper copyright collaboration.

Technically, research works~\cite{domainwatermark, zeromark} on dataset copyright protection and detection of unauthorized training data usage are progressing, which has positive implications for IGV in game development.

\begin{tcolorbox}[colframe=gray,colback=gray,coltext=white]
\textbf{Security Issues}: What measures can be implemented to prevent the generation of harmful content by generative models such as IGV?
\end{tcolorbox}

IGV systems can build on safety techniques developed for video generation, but their effectiveness must be assessed for sustained interactive use. Candidate measures include:
(1) Value alignment~\cite{safesora, safeld}, for example through RLHF during post-training, to reduce harmful outputs;
(2) Harmful-content detection~\cite{mllmasjudge, llavaguard} using VLMs to identify and filter problematic outputs, subject to the latency requirements of gameplay.
These measures are not guarantees: adversarial inputs, missed detections, false positives, and distribution shifts can cause failures. Bounded generation policies, ongoing testing, human oversight, and mechanisms to interrupt or fall back from unsafe generation remain important parts of deployment.

\begin{tcolorbox}[colframe=gray,colback=gray,coltext=white]
\textbf{Environmental Impact}: How should the energy costs of training and continuous generation be considered?
\end{tcolorbox}

Lower asset-production effort does not by itself imply lower environmental impact. Assessment should account for training and adaptation, runtime inference over actual play time, supporting services, and hardware use. Efficiency techniques and hybrid deployment may reduce these demands, but their net benefit depends on usage and should not be assumed from compute-price trends alone. Energy use belongs alongside monetary cost and player experience in evaluating whether IGV is appropriate for a deployment.

\begin{tcolorbox}[colframe=gray,colback=gray,coltext=white]
\textbf{Creativity Issues}: Can IGV serve as a creative tool, allowing for deep human creativity?
\end{tcolorbox}

We believe IGV can enhance human creativity for the following reasons:
(1) Interactive generative video technology eliminates mechanical and repetitive tasks in game development, such as debugging, writing basic code, and building standard scenes. By automating these uncreative aspects, developers can channel their energy into creative endeavors, focusing on innovative gameplay design and unique artistic expressions that truly matter to the gaming experience.
(2) IGV breaks down technical barriers, making game development accessible to creators from professional studios to independent developers. This democratization enables more diverse voices to enter the gaming industry, each bringing their unique perspectives and creative visions. Like other AIGC applications, it enables creators to realize their ideas without technical constraints, as demonstrated by artist Sofia Crespo's work\footnote{\url{https://en.wikipedia.org/wiki/Sofia_Crespo}} that blends technology with organic art, showing how AI amplifies creativity.

\begin{tcolorbox}[colframe=gray,colback=gray,coltext=white]
\textbf{Democratization Issues}: Does democratizing game creation diminish its value?
\end{tcolorbox}

We believe that democratizing game creation will not diminish its value, but rather enhance the overall value and creativity of the entire field. Here is our analysis and examples:

The democratization of gaming won't diminish creative value. The widespread availability of technology enables more people to enter this field, generating more diverse creative thinking and innovative designs. Creation value lies in innovation and personalization, not just technical difficulty. Through this technology, even ordinary users can create games with unique characteristics and personal style, which holds its own distinctive value.
A good example is the opening up of image generation models, which hasn't diminished the value of artistic creation. People with varying levels of professional expertise have shared numerous new artistic works on Civitai~\cite{civitai}, which has actually enhanced the creativity in this field and the value of its works.

\begin{tcolorbox}[colframe=gray,colback=gray,coltext=white]
\textbf{Labor Issues}: How should we view the potential negative impact of highly automated productivity tools like GGE on labor in the gaming industry?
\end{tcolorbox}

We acknowledge the labor impact concern with generative AI. IGV aims to enhance productivity and creativity rather than replace human workers. We advocate for measures like education and support programs to help industry professionals leverage AI tools, ensuring positive industry transformation. \finalrev{Concrete measures include industry-funded reskilling programs for artists and designers to become AI-directed content curators, the emergence of new roles (e.g., prompt engineers for game worlds, AI quality assurance specialists), and gradual adoption timelines that allow workforce adaptation rather than abrupt displacement.}

\begin{tcolorbox}[colframe=gray,colback=gray,coltext=white]
\textbf{Commercial Distribution and Content Rating}: How can generative games satisfy platform certification and content rating requirements?
\end{tcolorbox}

Current platform certification processes (e.g., Sony TRC, Nintendo Lotcheck, ESRB/PEGI ratings) assume content determinism: reviewers must be able to inspect all possible player experiences. Generative systems with unbounded output spaces fundamentally challenge this assumption. Similarly, regional publishing requirements (such as China's game license system) mandate pre-approval of fixed content. Possible mitigations include runtime content filters with certified safety models, pre-baked guardrails that bound the generation space, and a shift from auditing specific outputs to auditing model capabilities. We note this as a non-technical but structurally important barrier to IGV commercialization that will require collaboration between the AI and game industry communities.

\newpage
\section{One Possible Integration Workflow (Illustrative)}
We do not propose a normative workflow; the space of possible GGE-integrated pipelines is large and depends on game genre, studio size, and target platform. Below we sketch one illustrative example, intended as a concreteness aid rather than a prescription, elaborated here in response to prior community interest in what an end-to-end GGE development cycle might look like.

\begin{tcolorbox}[colframe=gray,colback=gray!10,sharp corners=southwest,boxrule=1pt,title={Concept Stage}]
 \textbullet\ \textbf{Concept Design}: Define core game elements (gameplay mechanics, story, target audience, art style) through LLM consultation and convert to IGV condition prompts.

\textbullet\ \textbf{Prototype Development}: Select suitable base models based on computing power and performance requirements, and develop prototypes using initial prompts for feasibility testing.
\end{tcolorbox}

\begin{tcolorbox}[colframe=gray,colback=gray!10,sharp corners=southwest,boxrule=1pt,title={Production Stage}]
\textbullet\ \textbf{Asset and Logic Requirements}: Create detailed prompts for specific assets and logic requirements (e.g., character model descriptions, area map sketches, level-up rule systems).

\textbullet\ \textbf{Training Data Collection and Model Fine-tuning}: Fine-tune models with targeted game data (e.g., collecting copyright-free space movie/game assets for a space exploration game).
\end{tcolorbox}

\begin{tcolorbox}[colframe=gray,colback=gray!10,sharp corners=southwest,boxrule=1pt,title={Testing Stage}]
\textbullet\ \textbf{Functionality Testing}: Test prompt-based content generation and screen for harmful content.

\textbullet\ \textbf{Compatibility and Performance Testing}: Optimize performance across different devices with necessary algorithm/hardware acceleration.
\end{tcolorbox}

\begin{tcolorbox}[colframe=gray,colback=gray!10,sharp corners=southwest,boxrule=1pt,title={Post-launch Stage}]
\textbullet\ \textbf{Content Updates}: Update model parameters and prompts for new content (DLCs, characters, events).

\textbullet\ \textbf{Data Analysis and Optimization}: Use player behavior data (with consent) for model fine-tuning and reinforcement learning.
\end{tcolorbox}

\begin{tcolorbox}[colframe=gray,colback=gray!10,sharp corners=southwest,boxrule=1pt,title={Enabling Conditions}]
The successful implementation of this workflow relies on these key factors:

\textbullet\ (1) \textbf{Model Capability}: Robust base IGV models that support efficient control, fine-tuning, and fast inference.

\textbullet\ (2) \textbf{Data Accessibility}: Well-established data sharing and copyright mechanisms that enable legal and cost-effective access to high-quality training data.

\textbullet\ (3) \textbf{Computing Resources}: Accessible AI computing infrastructure, either through local hardware resources or affordable cloud computing services.
\end{tcolorbox}

\noindent The above factors are enabling conditions, not guarantees; the actual shape of GGE-integrated workflows will emerge from iteration between engine vendors, studios, and the research community over the coming years.

\end{appendices}

\end{document}